\documentclass[twocolumn,10pt]{asme2ej}
\usepackage{booktabs}
\usepackage{multirow}
\usepackage{graphicx} 
\usepackage{subcaption}
\usepackage[table,xcdraw]{xcolor}
\usepackage{url}
\newcommand{\etc}{{\em etc.}}

\title{BIKED: A Dataset for Computational Bicycle Design with Machine Learning Benchmarks}

\author{Lyle Regenwetter
    \affiliation{
	Dept. of Mechanical Engineering\\
	Massachusetts Institute of Technology\\
	Cambridge, MA, 02139\\
    Email: regenwet@mit.edu
    }	
}

\author{Brent Curry
    \affiliation{
    BikeCAD.ca\\
    Ottawa, Ontario, Canada\\
    K1J 6E9\\
    Email: info@bikecad.ca
    }	
}

\author{Faez Ahmed
    \affiliation{
	Dept. of Mechanical Engineering\\
	Massachusetts Institute of Technology\\
	Cambridge, MA, 02139\\
    Email: faez@mit.edu
    }	
}

\begin{document}

\maketitle    

\begin{abstract}
{\it 
In this paper, we present ``BIKED,'' a dataset comprised of 4500 individually designed bicycle models sourced from hundreds of designers. We expect BIKED to enable a variety of data-driven design applications for bicycles and support the development of data-driven design methods. The dataset is comprised of a variety of design information including assembly images, component images, numerical design parameters, and class labels. In this paper, we first discuss the processing of the dataset, then highlight some prominent research questions that BIKED can help address. Of these questions, we further explore the following in detail: 1) How  can we explore, understand, and visualize the current design space of bicycles and utilize this information? We apply unsupervised embedding methods to study the design space and identify key takeaways from this analysis. 2) When designing bikes using algorithms, under what conditions can machines understand the design of a given bike? We train a multitude of classifiers to understand designs, then examine the behavior of these classifiers through confusion matrices and permutation-based interpretability analysis. 3) Can machines learn to synthesize new bicycle designs by studying existing ones? We test Variational Autoencoders on random generation, interpolation, and extrapolation tasks after training on BIKED data. The dataset and code are available at \url{http://decode.mit.edu/projects/biked/}
}
\end{abstract}

\setlength{\belowdisplayskip}{5pt} \setlength{\belowdisplayshortskip}{5pt}
\setlength{\abovedisplayskip}{5pt} \setlength{\abovedisplayshortskip}{5pt}


\section{Introduction}
Rapid advancements in machine learning research fields are often catalyzed by the introduction of quality publicly available datasets. Open source datasets often serve as benchmarks to evaluate the performance of different algorithms on a level playing field. In computer vision, for example, a set of well-known datasets (Imagenet~\cite{deng2009imagenet}, CIFAR-10~\cite{krizhevsky2009learning}, MNIST~\cite{lecun1998gradient}, \etc)  serve as the standard benchmarks for evaluating new techniques and methods and determining the state of the art. Data-driven design is a growing field that has increasingly tapped into machine learning and other data-driven methods to realize design goals, including automated design. New methods in data-driven design research are often tested on existing datasets like 3D model datasets and topology optimization (TO) datasets, however, these have drawbacks that limit the scope of their applications. We introduce the BIKED dataset with the primary intent of supporting new data-driven design methods both in general and in bicycle design specifically. In particular, we support the development of methods that can leverage detailed parametric design information, such as explicit dimensions and component categories.

Data-driven methods have the potential to enhance bicycle design exploration and generation, just as they have in other fields. We found the bicycle (bike) design problem to be an ideal domain for the testing and development of data-driven methods. Bicycles are machines with hierarchies of subcomponents, complex part dependencies, and large variations between designs. This variation largely stems from the broad set of use cases (road races, stunts, cargo transportation), physical considerations of the rider (body dimensions, flexibility, posture), and human preferences (style, aesthetics, budget). Due to this diverse set of constraints, bikes greatly benefit from customization. This makes the bicycle an excellent candidate for data-driven design, which typically leverages the design expertise contained in previous products, studies the diversity of existing designs, and captures the statistical trends of human preference. Compared to a manual design process, which is common in the bicycle industry, data-driven approaches can accelerate new bike design tasks and lead to novel and high-performing designs. 

Significant advancements to the bicycle design field also have the potential to positively impact society. As of 2015, there were estimated to be at least 580 million privately owned bikes worldwide~\cite{oke2015tracking}. Despite the continued popularity of bicycles, further increasing bicycle use and ownership can improve public health~\cite{oja2011health}, reduce traffic congestion~\cite{hamilton2018bicycle}, and positively impact the environment ~\cite{edenhofer2015climate}. We hope that by developing new data-driven bicycle design methods, advancements in the field of bicycle design can be realized and bicycles can subsequently be made more accessible to more people. This impact will be especially pronounced if data-driven design methods can enable increased customizability for particular user groups for whom bicycles may not presently be a competitive transportation option. 

BIKED---a dataset of bicycle models--is organized and curated specifically for data-driven design of bicycles\footnote{Dataset and code available at:  \url{http://decode.mit.edu/projects/biked/}}. Data is sourced from a rich archive of CAD files from the specialized BikeCAD software, which is primarily used by professional frame designers and bicycle enthusiasts. A sequential data curation process was carried out as detailed in Section~\ref{methodology} to extract parametric design data. As we demonstrate in Section~\ref{dimenredux}, this parametric data captures the rich variations between designs in the dataset and is highly revealing of many parameters of interest like bicycle type. 

While we present the dataset itself as the key contribution of this work, we also outline and explore several key research questions that BIKED may address: How can we explore, understand, and visualize the current design space of bicycles and utilize this information? When designing bikes using algorithms, under what conditions can machines understand the design of a given bike? Can machines learn to synthesize new bicycle designs by studying existing ones? We dedicate the latter portion of this paper to exploratory work on these research questions. 

In Section~\ref{dimenredux} we consider the question of how we can explore, understand, and visualize the current design space of bicycles and furthermore utilize this information. We approach this problem through unsupervised embedding methods.  These methods can generate easy-to-visualize low-dimensional maps of the design space. By studying the distribution of bicycles across the space, especially with bicycle class annotations, we can understand trends in the data such as what classes tend to overlap or distance from one another. By identifying sparse regions in the design space, we can potentially even identify design opportunities that are underrepresented in the current bicycle industry.

In Section~\ref{classification} we explore the conditions necessary for machines to understand the design of a given bike. We formulate a classification problem to evaluate this question. Specifically, we investigate what quantity of data, type of data, and algorithm yields high classification performance. To do so, we evaluate and provide baseline results for 12 classification algorithms on several training set sizes. We then construct, tune, and train three deep neural networks to predict class results based on different types of input data provided within BIKED. These neural networks provide a baseline for higher performance models and yield valuable insight into the predictive information contained within the parametric and image data. In addition to maximizing performance, practitioners must understand the patterns identified by machine learning. We examine the behavior of the classifiers by generating and analyzing confusion matrices. Furthermore, to understand what design attributes are important for each bike style, we pose this problem as an interpretability analysis problem for a classification model. Using Shapley Additive Explanations~\cite{NIPS2017_7062}, we explore how individual parameters impact class predictions and discuss the design implications of our results. 

Finally, in Section~\ref{synthesis}, we address the last of our three big questions: When designing bikes using algorithms, can machines create a unique bicycle design that they have not seen before? We present applications and baseline results of two Variational Autoencoders (VAEs) applied to the parametric and image data respectively. VAEs support a host of applications including dimensionality reduction and novel data synthesis. We explore several types of data synthesis: random sampling (exploring the design space), interpolation (synthesizing new designs between two existing designs), and extrapolation (synthesizing new designs away from an existing design) in both the original parameter space and the latent space of the VAEs. 

Inspiring other researchers to use standardized problems and datasets for data-driven design applications and algorithm development is one of the key goals of BIKED. To that end, we provide a suite of algorithms (Variational Autoencoders, convolutional neural networks, Gaussian processes, ensemble classifiers, \etc{}) and their performance values throughout our exploration of the aforementioned research questions.

\section{Background and Previous Work}
In this section, we discuss the background and related work in bicycle design optimization, compare BIKED to other datasets commonly used for data-driven design, and discuss the BikeCAD software and design archive. We also note that this paper is primarily an expansion of previous work~\cite{regenwetter2021biked}. 

\subsection{Bicycle Design and Optimization}
Bicycle design and optimization is a well-researched field. Numerous treatises have explored principles of bicycle design in the centuries since the earliest predecessors to the modern bicycle~\cite{wilsonschmidt2020, Sharp1977}. Today, significant research effort is dedicated to improving bicycle aerodynamics~\cite{chowdhury2011, chowdhury2012, malizia2020bicycle} and structure~\cite{suppapitnarm2004conceptual}. Other studies explore practices of bicycle sizing and fitting~\cite{laios2010ergonomic, swart2019cycling}. While some make use of the wide availability of anthropometric data, many are limited by the availability of bicycle design data. We expect that a quality dataset of bike designs including comprehensive parametric information will enhance future simulation-based bicycle design studies. 

\subsection{Comparison to Other Datasets}
Research in this domain of data-driven design often taps into 3D model datasets like Shapenet~\cite{chang2015shapenet} and Princeton ModelNet~\cite{wu20153d}. These datasets are particularly useful in 3D shape synthesis applications. Though some of this shape synthesis research considers adherence to physical laws, the nature of these datasets tends to emphasize research focused on visual fidelity rather than physical function~\cite{yang2020dsm, gao2019sdm, mo2019structurenet, li2017grass}. In contrast, BIKED contains thousands of design parameters that explicitly specify design information ranging from local component geometry to overall design layout. Certain design information can be extracted from 3D models, but the level of detail falls short of the comprehensive numerical design parameters that BIKED provides.

Data-driven design research frequently uses Topology Optimization (TO)-based approaches, which often depend on TO datasets. While TO datasets can provide extensive information for models learning to optimize local geometry~\cite{sosnovik2019neural, zhang2019deep}, approaching full design synthesis tasks using TO is often impractical, and generalizing TO datasets to new design domains poses an additional challenge. As such, BIKED enables methods and functionality that would not be feasible to develop using TO datasets. 

\subsection{BikeCAD} \label{bikecad}
\begin{figure}[h]
    \centering
    \includegraphics[width=0.39\textwidth]{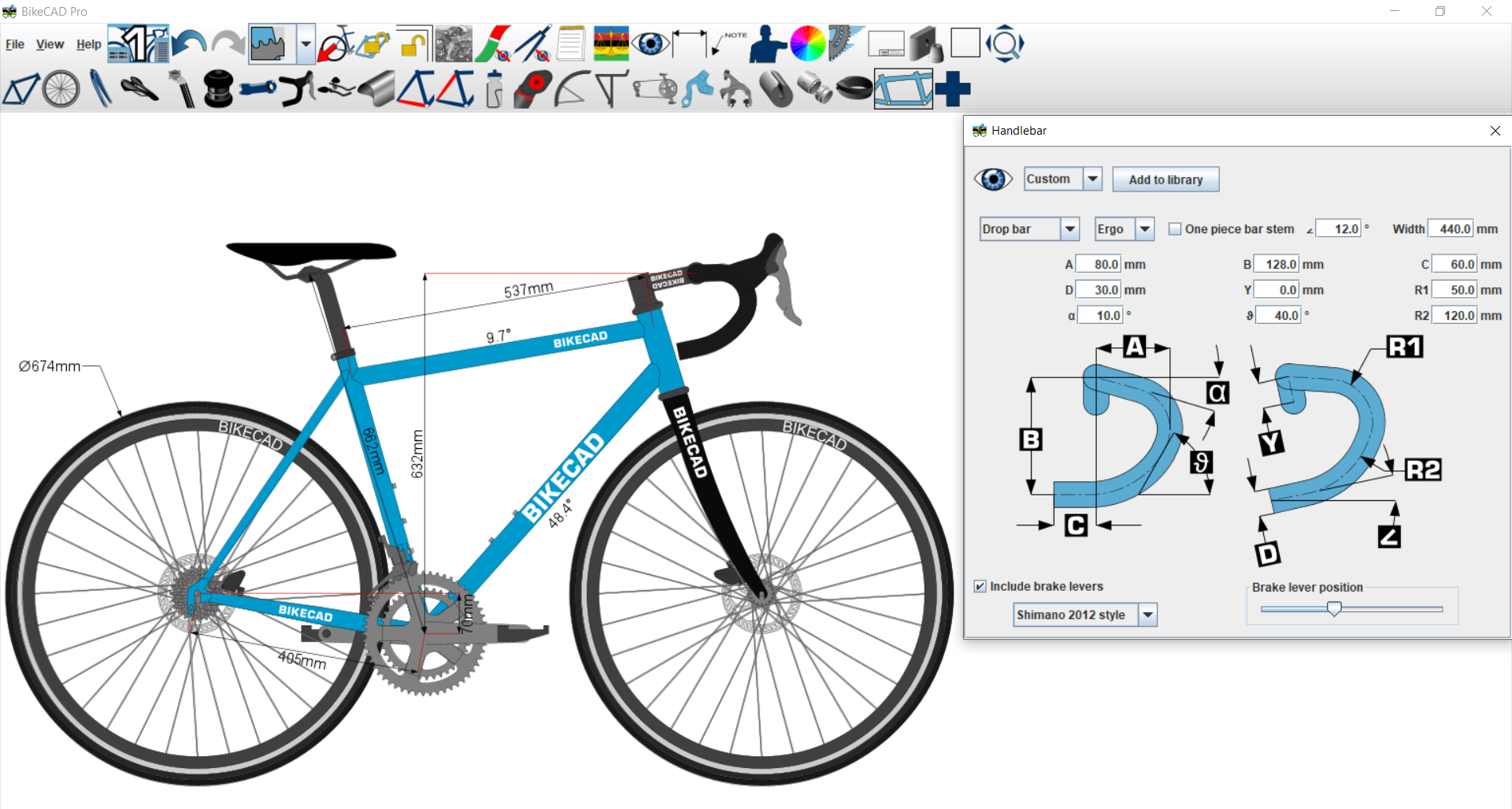}
    \caption{Screenshot from BikeCAD software showing active model with some dimensional labels and an open menu.}
    \label{fig:bcad}
\end{figure}

\begin{figure}[h]
    \centering
    \includegraphics[scale=.35]{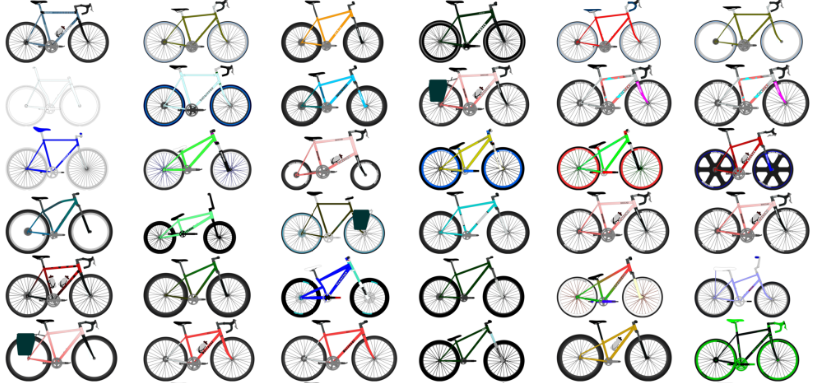}
    \caption{Tile images of sample bike designs. Backgrounds and dimensional labels are removed for visibility.}
    \label{fig:tile}
\end{figure}

BikeCAD is a parametric computer-aided design (CAD) software optimized for bicycle design. It features a live-updating model and numerous design menus that help users customize bicycle geometry and features, as shown in Figure~\ref{fig:bcad}. The BikeCAD website features an open archive of user-submitted BikeCAD designs~\cite{curry_2018_archive}. As of July 2020, the archive had accrued roughly 5000 bicycle designs since its inception in 2011. Figure~\ref{fig:tile} shows tile images of several randomly sampled bikes from the dataset. Since designs span many iterations of the BikeCAD software, the advancement of the program over the years has enabled increased complexity in more recent designs on the archive. Each entry in the dataset contains the BikeCAD file, an image, a rating out of 5, the BikeCAD version used to create the design, and a design identification number (ID). Additionally, some designs in the archive contain the ID that they are based on and several parametric dimensions. Of this information on the archive, only the BikeCAD files themselves were used in BIKED. Rating information was not included since most designs are unrated and few designs have more than a single rating. Additionally, though tracking the design IDs that each design is based on would make for an interesting graph of design progression, many design ID links are no longer functional since their referenced designs have been removed from the archive. 

Since BikeCAD's primary user base is comprised of frame builders and bicycle enthusiasts, many designs are high-quality and are furthermore self-selected as showcase designs. We also note that the BikeCAD software promotes quality through realistic default values, comprehensive design and analysis tools such as toe overlap and lean angle analysis, as well as a suite of tools to check the design's adherence to Union Cycliste Internationale (UCI) standards. All in all, this combination of factors suggests a reasonably high quality for the majority of dataset designs.

\section{Methodology} \label{methodology}
\begin{figure}[h]
    \centering
    \includegraphics[scale=.54]{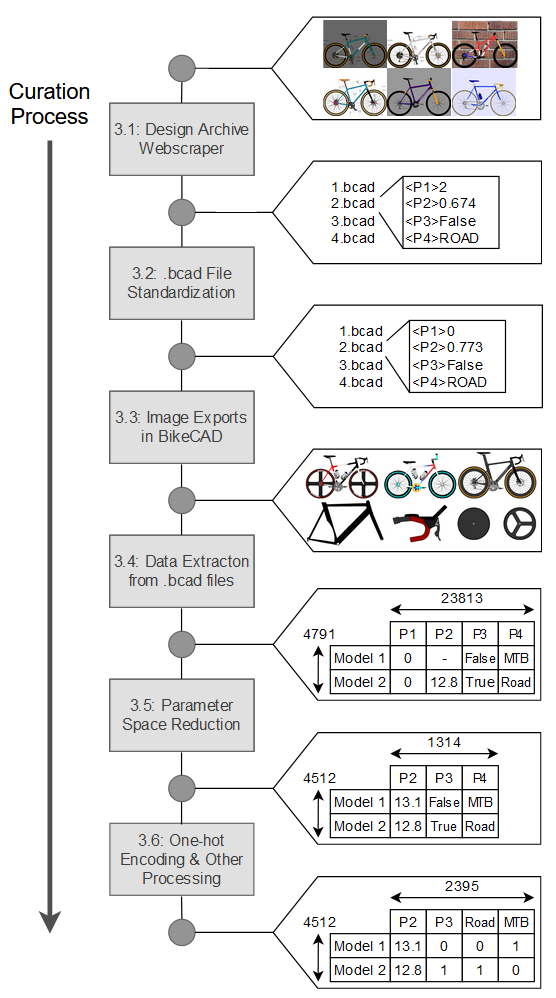}
    \caption{Flow diagram of dataset curation process. Processing steps (gray boxes) are labeled according to the section in which they are discussed. Data snapshots (right) demonstrate sample data generated by the previous processing step.}
    \label{fig:flow}
\end{figure}

In this section, we discuss the various steps taken to generate and process the dataset. An overall flow diagram of the data curation progression is shown in Figure~\ref{fig:flow} and the following labeled sections correspond to the processing operations listed.

\subsection{Sourcing from Design Archive}
To process the dataset locally, we downloaded designs from the BikeCAD design archive. We were able to download 4800 of the designs posted on the archive as of July 2020. Original ordering on the design archive was preserved in the subsequent processing of the data.

\subsection{File Standardization in BikeCAD}
BikeCAD files are XML files structured as a collection of data parameters with corresponding values. Certain parameters have different meanings depending on flags in other parts of the file. To ensure that the parameters in the files referred to the same geometric dimensions across different designs, BikeCAD's internal conversion formulas were applied using a custom version of the BikeCAD software developed by one of the authors. 9 files were corrupt or unprocessable leaving 4791 successfully standardized designs. 

\subsection{Bicycle Assembly and Component Image Exports}
\begin{figure}
    \centering
    \begin{subfigure}[b]{0.2\textwidth} 
        \centering
        \includegraphics[width=\textwidth]{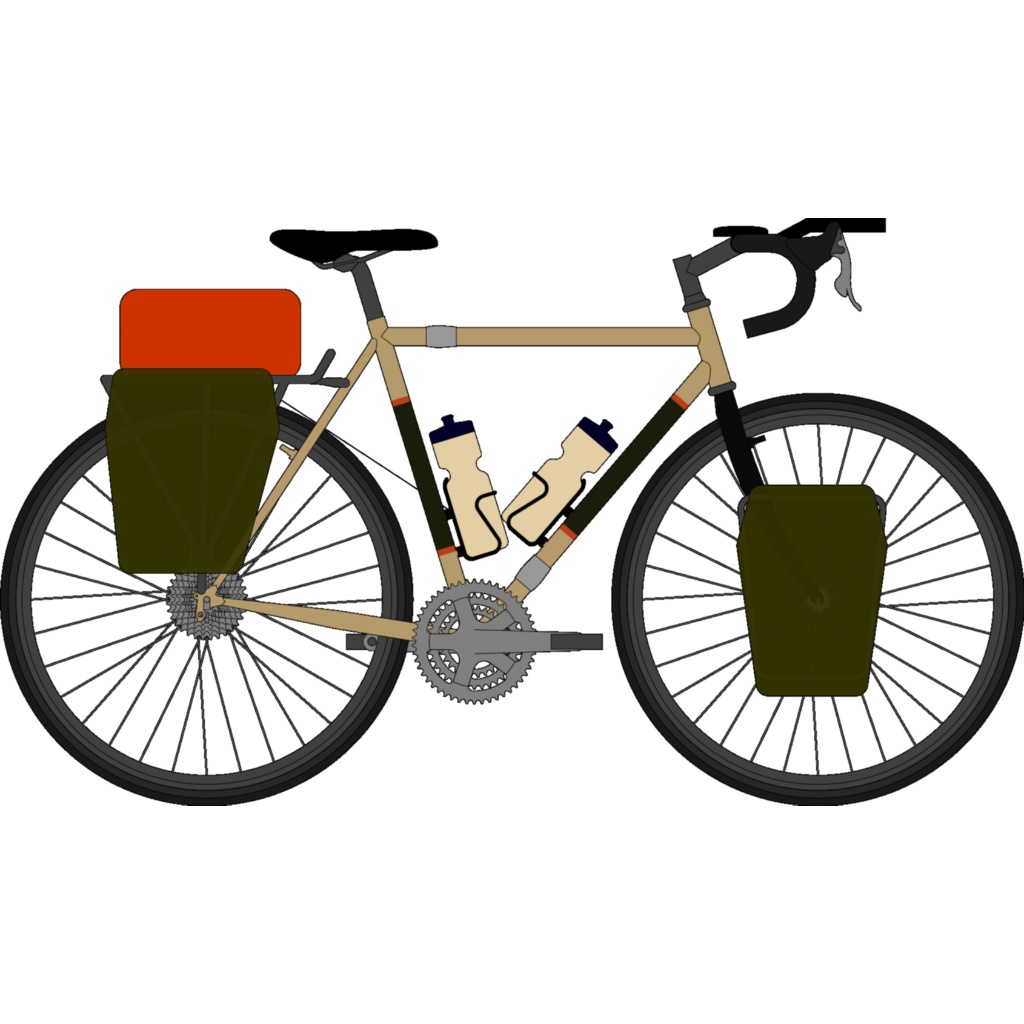}
        \caption[segbike]%
        {{\small Full Bike}}    
        \label{fig:segbike}
    \end{subfigure}
    \hfill
    \begin{subfigure}[b]{0.2\textwidth}  
        \centering 
        \includegraphics[width=\textwidth]{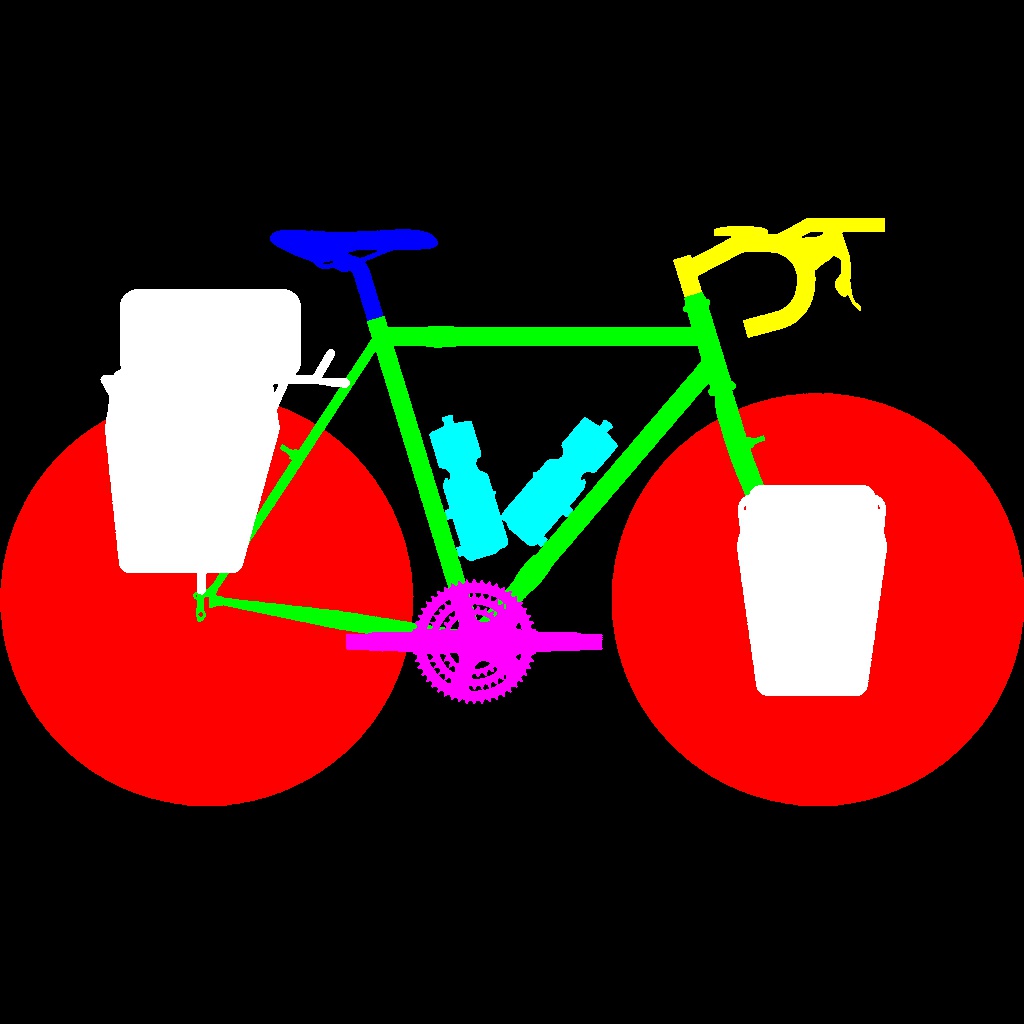}
        \caption[Segmentation]%
        {{\small Image Segmentation}}    
        \label{fig:segmentationoverlay}
    \end{subfigure}
    \vskip\baselineskip
    \begin{subfigure}[b]{0.4\textwidth}
        \centering
        \includegraphics[width=\textwidth]{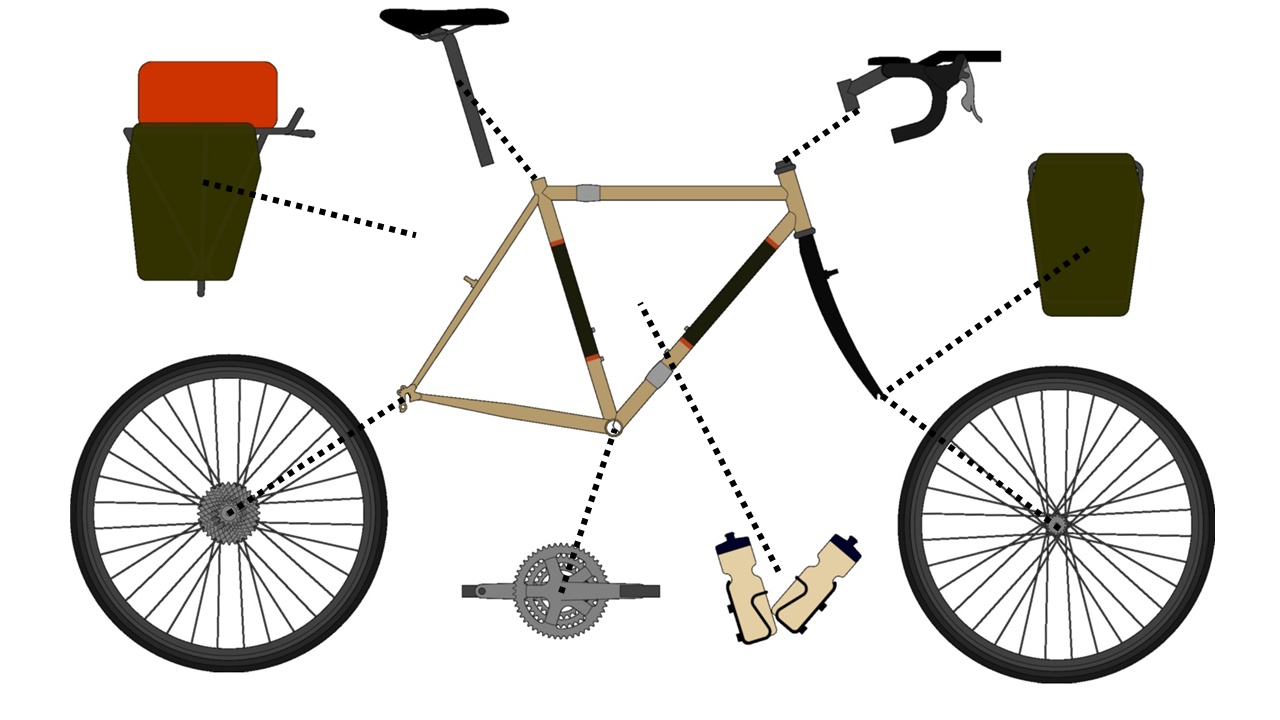}
        \caption[Exploded View]%
        {{\small Exploded View}}    
        \label{fig:explode}
    \end{subfigure}
    \caption[Components]
    {\small Segmentation of a bicycle into 7 component parts. Figure ~\ref{fig:segbike} shows the bike assembly image and Figure ~\ref{fig:segmentationoverlay} shows a semantic mask of its segmentation into components. Figure ~\ref{fig:explode} demonstrates an exploded view showing the 7 component groups. Observe that portions of the bike that were covered in the original image are present in the component images, such as the portion of the wheels that were covered by the stem, chainstays, and racks. Exploded views are not included in the dataset.} 
    \label{fig:segmentation}
\end{figure}

To remove dimensioning lines and backgrounds from the bike models' corresponding images, we directly edited label visibility and set plain backgrounds in the BikeCAD files, then exported these clean bike images. Next, segmented component images were rendered for 5 essential components (frame, saddle, handlebars, wheels, cranks) and two nonessential components (cargo racks, bottles) that only appeared in some designs. BikeCAD models are structured as a collection of component parts that form an assembly. Each component image was rendered and exported from BikeCAD after editing the BikeCAD files to hide all other components. 4510 bikes were compatible with this process, yielding 5 sets of 4510 component images (essential components), one set of 965 (bottles), and one set of 346 (racks). An example segmentation on one bike design is demonstrated in Figure ~\ref{fig:segmentation}. The components in the segmented images maintain the same position and scale as the original assembly images. In this sense, the segmented images can be interpreted as a ``layering'' of the full assembly image into bicycle component image slices. Semantic masks of the segmentation are also included.

\subsection{Data Extraction from BikeCAD Files}
For ease of processing, we parsed the individual standardized BikeCAD files and compiled all parameter data in a tabular data structure. In total, 23818 unique parameters (or features) were collected from the files, though many parameters were quite sparse across the design space since certain designs had no entries for many parameters.

\subsection{Parameter Space Reduction} \label{paramredux}
Lowering the dimensionality of a dataset can help reduce computational cost and memory requirements when applying data-driven models. Given that our original data was 23818 dimensional, dimensionality reduction was critical for efficient data-driven analysis. To reduce the dimensionality of the parameter space, several parameters were removed from consideration. First, parameters with no bearing on bicycle function such as color, text, and positioning of dimensional labels were identified and dropped from the data. Next, parameters defining additional non-standard tubes, water bottles, or tandem features were removed, and any bike with a data value under one of these fields was dropped. This removed some unique designs but drastically reduced the dimensionality of the data by dropping sparsely inhabited parameters. Additionally, each design was manually considered and any designs that were not bikes (scooters, motorcycles, wheelchairs, \etc{}) were dropped. Finally, any parameters that were empty for every design or for which every design had the same value were dropped. After these steps, 4512 bikes of 4791 and 1314 parameters of 23813 remained, resulting in a 94.5\% reduction in dimensionality. We note that these 4512 designs constitute a different subset of the original 4800 than the 4510 designs compatible with the segmentation process. 

Approximately $5\%$ of designs lacked one or more essential components, which in practice implied that the designer turned off the visibility of said components in the software. For example, some designs originally had only a frame visible, presumably implying the designer was only intending to design a frame. Since we did not want incomplete designs in the dataset, critical components of bike designs were used in the processing of the dataset regardless of whether they were hidden or not. In other words, visibility parameters of critical components were dropped from the parameter space. In cases like the standalone bike frame designs, it was unclear if the hidden components like wheels, saddle, and handlebars received any design consideration, but were included nonetheless. As such, some designs may include components not intended to be part of a final design. In contrast, non-essential features like bottle holders and fenders were included or excluded in the parameter space based on the visibility in the original design. We acknowledge that manually overriding component visibility of critical design components could have unintentionally included components that were not intended to be part of the designer's vision, however, the number of affected designs was minimal. In these rare cases, the bicycle model could have one or more components whose parameters were not adjusted from the defaults or were set in an earlier draft of the model. 

\subsection{Other Processing Steps} \label{otherprocessing}
Following the parameter space reduction, the parameter space consisted of four types of data: continuous, discrete, boolean, and categorical. Continuous and discrete parameters reflect some ordinal significance (i.e. similar numerical values imply actual real-world similarity in that parameter). In contrast, categorical values such as numerical class labels imply no such ordinal significance even if represented numerically. Some example parameters of each type are included in Appendix A1, and we refer the reader to the project page for a full list. Most data types in the parameter space could be easily classified (i.e. floating-point numbers are continuous). However, integer-valued parameters had to be individually considered and manually classified as categorical or discrete. Additionally, certain variables that would ideally be continuous had to instead be treated as categorical. For example, the bike ``size'' parameter was treated as categorical since it contained a medley of words and measurements with varied or ambiguous units (``xxl'',``623EET'',``Huge'',``52cm'',``26'', \etc{}). Since many standard machine learning methods do not support categorical data, we applied a commonly used method called one-hot encoding, which converts a categorical variable with $n$ unique categories into a set of $n$ boolean parameters. These steps left us with the distribution of variables shown in Table~\ref{tab:datatypes}.

\begin{table}[ht]
\caption{Assortment of types of parameters present in dataset before and after one-hot encoding}
\renewcommand{\arraystretch}{0.8}
\begin{tabular}{lllll}
&\multicolumn{2}{c}{Original}&\multicolumn{2}{c}{One-Hot}\\
\toprule
\textbf{Parameter Type}& \textbf{Num.}    & \textbf{Perc.}   & \textbf{Num.}  & \textbf{Perc.}  \\
\midrule
Cont. Variables     & 635  & 48.3\% & 635 & 26.5\%\\
Discrete Variables   & 424  & 32.3\% & 424 & 17.7\% \\
Booleans & 148  & 11.3\% & 1336 & 55.8\% \\
Catg. Variables & 107 & 8.1\%  & 0 & 0\%\\
\bottomrule
\end{tabular}
\label{tab:datatypes}
\end{table}

Next, missing and unknown values were imputed using k-nearest-neighbors imputation with $k=5$ neighbors.\footnote{Other imputation strategies such as mean imputation and median imputation are also provided as options in the associated code, but a detailed investigation contrasting imputation methods and hyperparameters was not performed in this study.} Values with a magnitude above a selected threshold were also discarded and then considered to be missing, as they were likely caused by corrupted data or data type issues. A cutoff magnitude of 100,000 was selected since most units were in millimeters or inches, and bikes were unlikely to be designed with component dimensions over 100 meters. Other outlier removal methods such as z-score cutoffs were examined and found to exclude quality designs without any discernible corrupted data. The final imputed data is the version of the data used in the discussion and sample applications discussed henceforth. 

\subsection{Image Regeneration} \label{Imregen}
\begin{figure}[h]
    \centering
    \includegraphics[scale=.22]{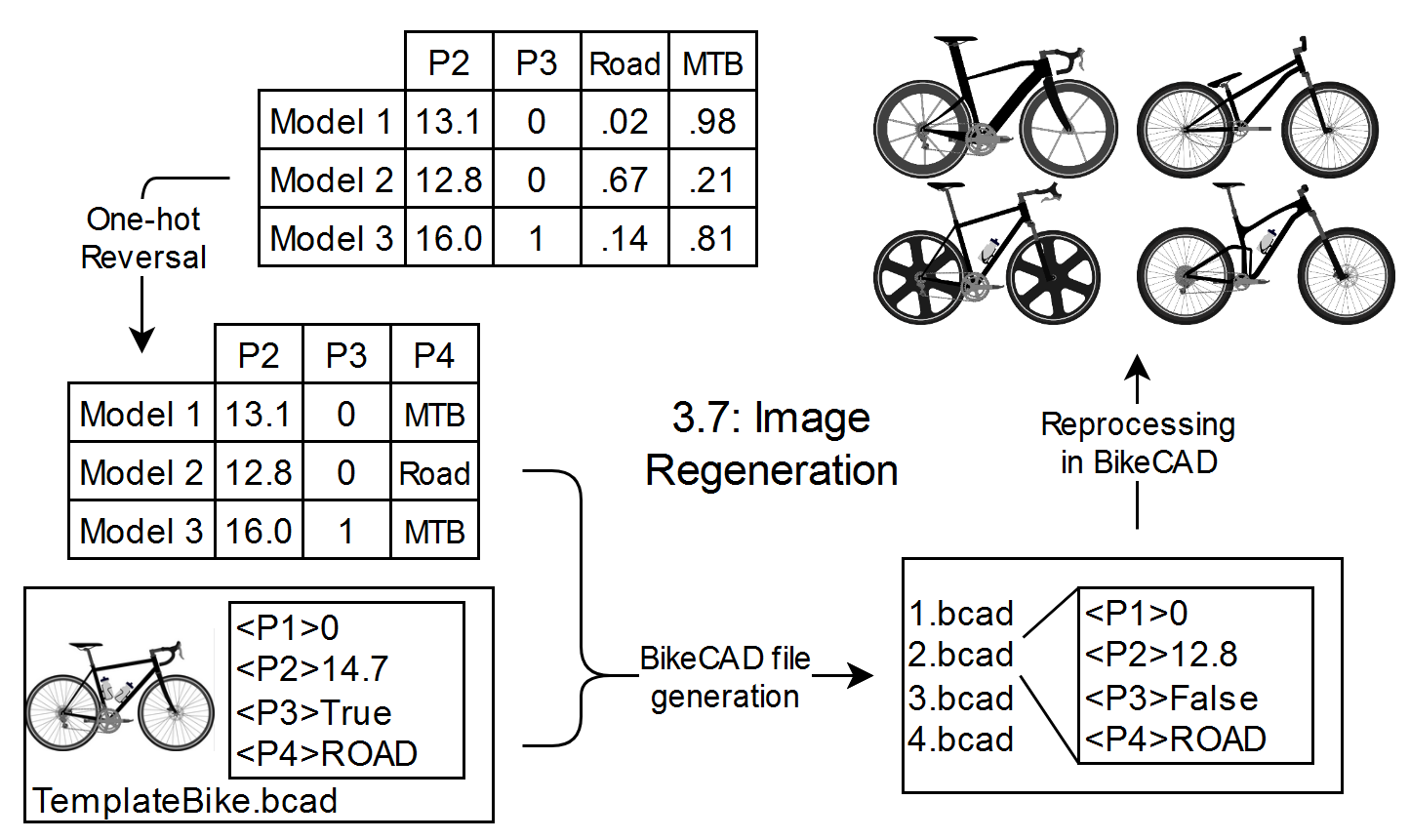}
    \caption{Process to generate images from parametric data}
    \label{fig:genflow}
\end{figure}

Because parametric data is difficult to quickly interpret, we developed a method to regenerate images from parametric data, as shown in Figure~\ref{fig:genflow}. We use this method to generate corresponding images from the fully processed parameter data. This method also applies to newly generated parametric data, which can assist in visualizing interpolation or novel bike generation results as we demonstrate in Section~\ref{synthesis}. In this process, the one-hot encoding is reversed by taking the most probable category to be the absolute truth. In the original dataset, one-hot vectors are populated with boolean values, but the method also supports probabilistic values to better accommodate generative methods. Next, bike data is inserted into a BikeCAD file template to generate new BikeCAD files from the standardized parametric data. Any fields from the template file that are present in the parametric data are overwritten and any absent fields are left at default values. These BikeCAD files are then opened in the software to export corresponding images. Since certain features like component colorings were dropped as described in Section~\ref{paramredux}, this regeneration process yields images with standardized colors as dictated by the template file. 

\subsection{Dataset Feature Summary} \label{Featuresum}
The data curation process yields several types of data at various levels of processing. The fully processed data consists of tabular parametric data after the processing steps detailed above and standardized images regenerated from curated parametric data. We also provide minimally processed data to support unforeseen applications that may require specialized curation steps. This data consists of raw tabular parametric data sourced directly from BikeCAD files with no processing and original bicycle images after background and dimensional label removal. Finally, we provide segmented bicycle component images to support component-wise learning tasks on images. We also note that the tabular data contains several parameters that can be used as labels for different learning tasks, such as bike class, model year, or individual component types. Note that these labels are neither rigorous nor standardized since all values are reported under the individual designers' discretion. For instructions on usage, we again refer the reader to the BIKED's project page, which links to a comprehensive usage guide.

\section{Data-driven Design Applications} \label{applications}
In this section, we consider three main research questions: How can we explore, understand, and visualize the current design space of bicycles and utilize this information? When designing bikes using algorithms, under what conditions can machines understand the design of a given bike? Can machines learn to synthesize new bicycle designs by studying existing ones? We address these research questions one by one in the following three subsections. While addressing these questions, we simultaneously hope to accomplish three goals:
\begin{enumerate}
    \item Provide a starting point and inspiration for researchers interested in using BIKED to expand upon one of these research questions or pursue new research directions supported by BIKED.
    \item Give baseline performance values for a variety of common data-driven tasks.
    \item Uncover patterns, relationships, and trends within the various forms of data included in BIKED.
\end{enumerate}
Discussion of other research questions that BIKED may help address are discussed in Section~\ref{future}. 

\subsection{Bicycle Design Space Exploration and Visualization} \label{dimenredux}
Extensive design data can open doors to better understand, visualize, and explore a design domain. We consider methods to uncover these insights and discuss how to furthermore utilize this information in making design decisions. By visualizing the relationships between different classes of bicycles, designers can identify trends, patterns, and potential gaps in the design space. To meet a broader set of customer needs, new classes like hybrid bikes are often introduced. Understanding what new classes can be introduced in the market or what features from one class can be incorporated in another class of bicycle can help designers efficiently target underrepresented segments of the industry.

To explore and visualize the bicycle design space, we use unsupervised embedding algorithms on BIKED. Design space embeddings can allow researchers to visually identify groups of similar designs in a dataset or identify areas of the design space that are sparse or completely empty. Design space embeddings can also be helpful for designers to place their own designs within the context of existing ones.

\paragraph{Bicycle Class Labels:}

To assist in visualization, we label designs using the bicycle ``style'' parameter, which we refer to as the bicycle ``class.'' Table~\ref{tab:classtypes} shows the distribution of bicycle classes in the dataset. We observe that road bikes are by far the most prevalent class (40.56\%) as labeled by the designers. We note that using bicycle ``styles'' as class labels has several drawbacks. Labels are reported by designers and are subjective and unstandardized. Classes are also not mutually exclusive: bike designs may potentially fit into more than one class, and several of the classes are conventionally considered subclasses of others (e.g. cyclocross, time trial, and touring bikes are subclasses of road bikes) \cite{century}. Despite this fact, each design in BIKED is assigned to exactly one class. Furthermore, several of the classes, like ``hybrid'' or ``other'' are not necessarily descriptive of an individual class of bike. Discovering new ways to represent bikes at a high level besides through mutually exclusive classes is a potential research direction that BIKED may support. 
\begin{table}[ht]
\centering
\caption{Assortment of bike classes present in the processed dataset. Mountain bikes are abbreviated as "MTB." Bikes classified as ``OTHER'' are labelled as such by the designer. The ``Remaining Label Categories'' group contains the remaining 10 explicit categories in descending order of prevalence: BMX, CITY, COMMUTER, CRUISER, HYBRID, TRIALS, CARGO, GRAVEL, CHILDRENS, FAT.}
\renewcommand{\arraystretch}{0.8}
\begin{tabular}{lll}
\toprule
\textbf{Bike Class} & \textbf{Count} & \textbf{Percent} \\
\midrule
ROAD         & 1865  & 41.33 \\
MTB          & 616   & 13.65 \\
TRACK        & 470   & 10.42 \\
OTHER        & 315   & 6.98 \\
DIRT\_JUMP   & 293   & 6.49 \\
TOURING      & 201   & 4.45 \\
CYCLOCROSS   & 151   & 3.35 \\
POLO        & 128   & 2.84 \\
TIMETRIAL    & 89    & 1.97 \\
Remaining Label Categories & 384   & 8.51\\
\bottomrule
\end{tabular}
\label{tab:classtypes}
\end{table}

\paragraph{ T-SNE embedding of full dataset:}

\begin{figure*}[h]
    \centering
    \includegraphics[scale=.35]{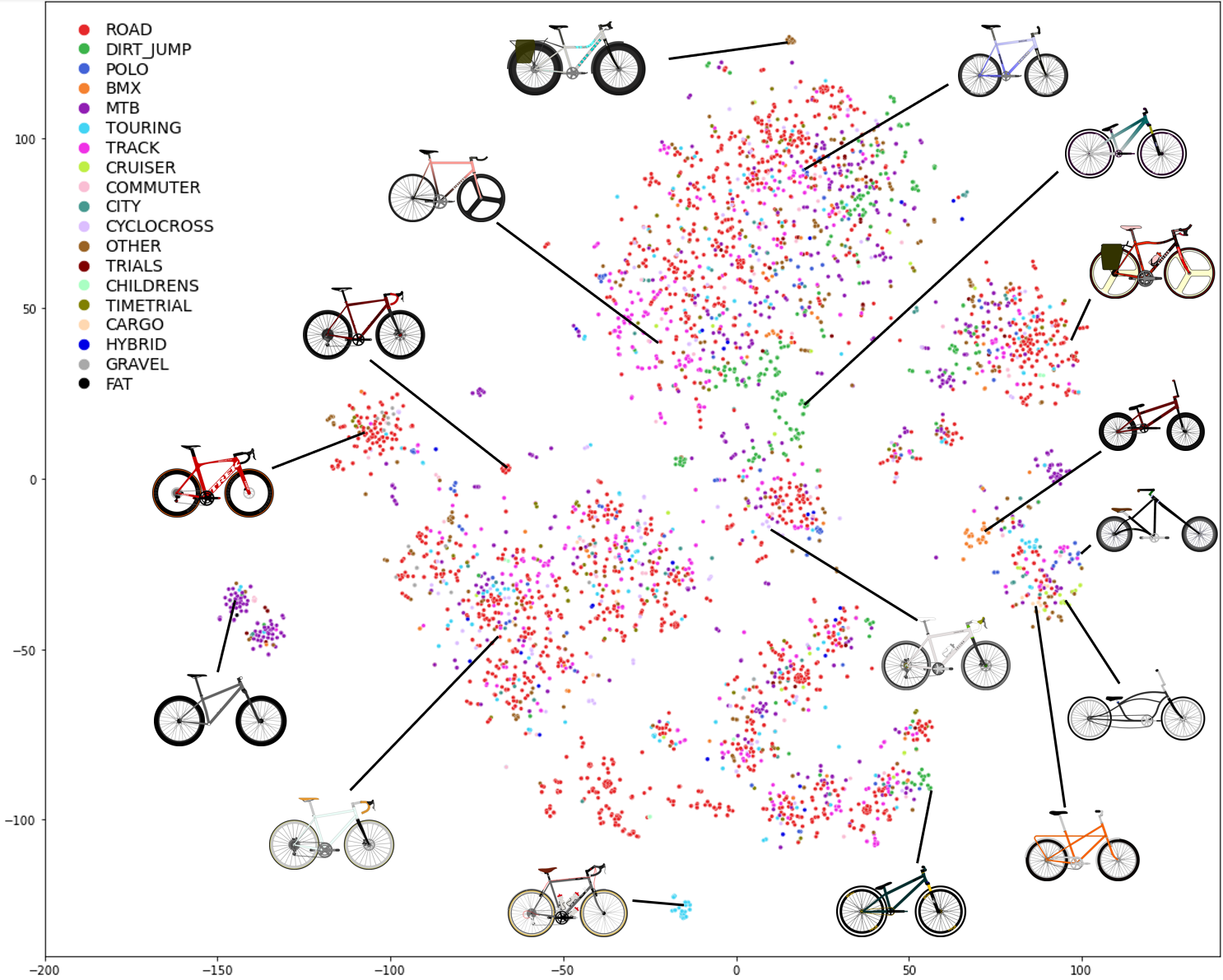}
    \caption{Visualization of the 2-dimensional embedding generated through t-Distributed Stochastic Neighbor embedding of parametric data. Horizontal and vertical axes denote the two embedding dimensions.}
    \label{fig:tsnelabeled}
\end{figure*}

To visualize the designs, we apply a t-Distributed Stochastic Neighbor Embedding (t-SNE) method~\cite{van2008visualizing} to the parameter space of the processed dataset. t-SNE is a dimensionality reduction algorithm that projects samples onto a lower-dimensional embedding space while keeping similar designs together and distancing dissimilar ones. An annotated plot of such an embedding is included in Figure~\ref{fig:tsnelabeled}. The tendency of classes to overlap illustrates the aforementioned non-mutually exclusive property of the bike classes and this embedding helps reveal the extent of the overlap and affected classes. We can also identify several unique clusters of bike designs.


\paragraph{PCA embedding of median designs:}

\begin{figure}[h]
    \centering
    \includegraphics[scale=.26]{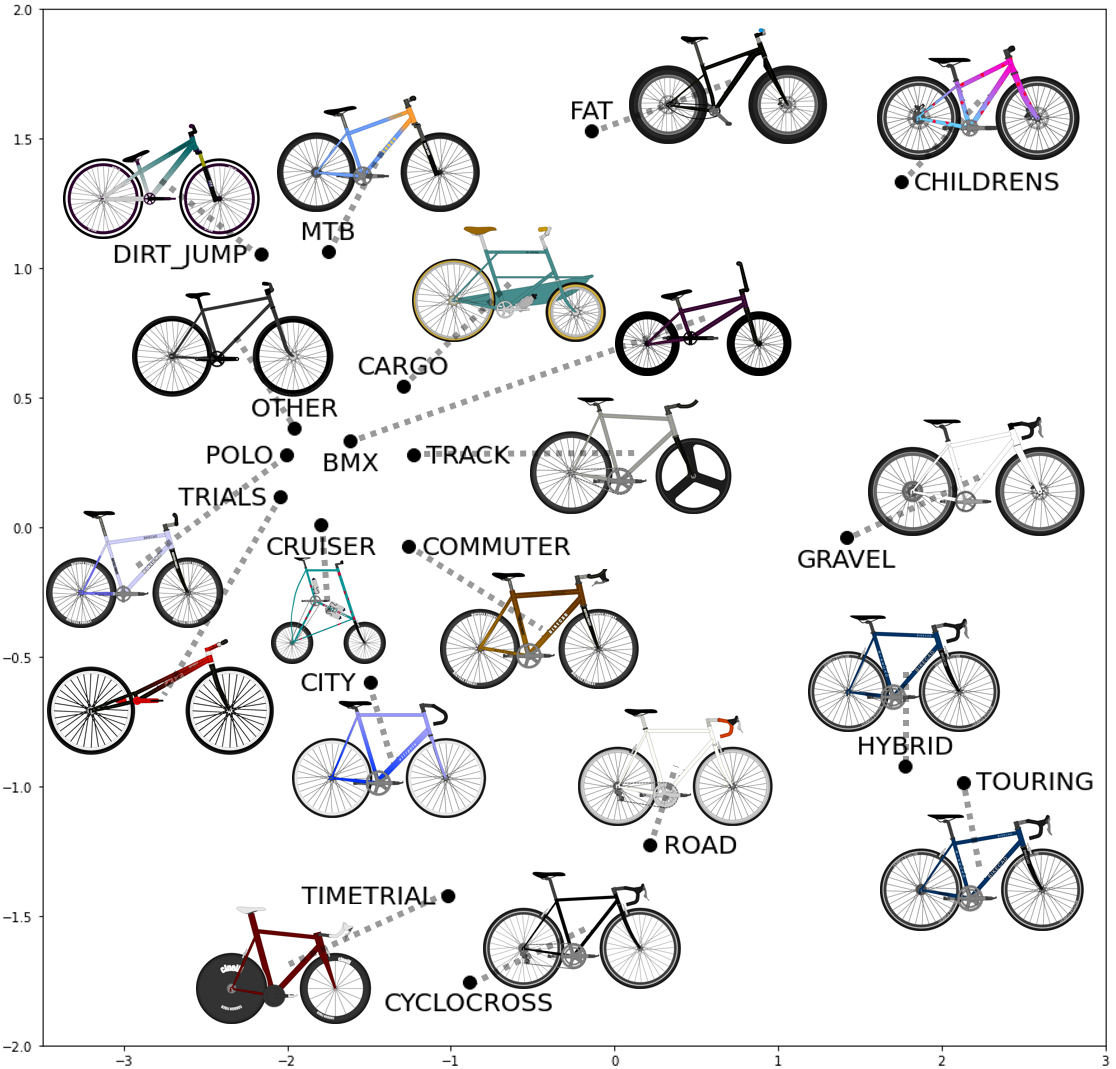}
    \caption{Visualization of ``median'' bikes of each class in a 2-dimensional Principal Component Analysis embedding generated using the 50 most significant parameters. The principle PCA component is shown on the horizontal axis and secondary component is shown on the vertical axis. }
    \label{fig:tsne}
\end{figure}

Next, we consider a more targeted approach by eliminating all parameters except the 50 deemed to be most impactful on bicycle classification. These specific parameters were determined using a SHAP analysis as discussed in Section~\ref{classification}. To better visualize the relationship between classes of bikes, we plot only the ``median'' bike of each class as a representative. We determine this ``median'' bike by applying min-max normalization to every parameter, taking the median of each normalized parameter over all bikes of that class, and finding the nearest design to this median value by euclidean distance. The results of a Principal Component Analysis (PCA) under these conditions are included in Figure~\ref{fig:tsne}. In this plot, we can observe several trends. For example, bikes with components and structure more similar to a conventional road bike tend towards the bottom right corner. Bikes with large wheels relative to absolute size tend toward the top.

We have demonstrated 2-Dimensional embedding examples that are particularly conducive to design space visualization, however, for tasks like identifying sparse regions in the design space, higher dimensional PCA embeddings may be more effective. In combination with the generative process described in Section 3.7, PCA can be a powerful tool. Designers can manually select points of interest in the embedding space, decode latent vectors, then directly generate full bicycle CAD models and images to evaluate these designs. We evaluate a similar generative process in Section 4.3 using the latent space of Variational Autoencoders, albeit using automated sampling instead of manual selection.

\subsection{Bicycle Classification} \label{classification}
To create meaningful designs, generative machine learning models must identify key characteristics which define a bicycle model and accurately predict which real-world class of design best describes it. In this section, we explore the conditions under which machines can understand the design of a previously unseen bike. Moreover, we examine how intelligent algorithms selectively process data to gain this understanding. To address these questions, we design a classification problem, experiment with various methods, document their performance, and analyze their behavior. In particular, we seek to understand how the algorithm, type of data, and quantity of data impact classification performance, as well as understand the behavior of high-performing classifiers. 

\paragraph{Baseline Classification Performance:} In the first part of this study, we explore how algorithm selection and dataset size impact classification performance on BIKED's parametric data. Table~\ref{tab:classification} presents an excerpt of a study contrasting 12 untuned models and 5 train set sizes, with each combination instantiated 10 times. We report top and average performance for four classification metrics: Accuracy, precision, recall, and F1. Full results and further details about this study are included in Appendix B1. 

We find that certain methods drastically outperform others in the classification task. Overall, the smaller neural network (3 layers) performs the best in the F1 metric and the Gaussian Process Classifier (GPC) performs best in overall accuracy. We note that the success of the GPC comes at the cost of significantly higher computational expense and this method would not be viable on a significantly larger dataset. Overall, the classification performance may seem underwhelming, but we discuss the challenges that come with this classification formulation later in this section. 

Another key takeaway is the improvement of both accuracy and F1 score with increased dataset size, a trend that remains clear into training sizes on the order of thousands. This suggests that additional bicycle designs would allow machines to better understand the bicycle design space and continue to provide valuable information to classification models. As such, a future expanded dataset with new BikeCAD archive designs would likely be valuable. 

\begin{table}
\renewcommand{\arraystretch}{1}
\centering
\caption{Classification accuracy for various classification models using differing quantities of training data points. Average scores are determined over 10 instantiations.}
\label{tab:classification}
\resizebox{\linewidth}{!}{%
\begin{tabular}{|l|lllll|}
\hline
\multicolumn{1}{|c|}{\multirow{2}{*}{Model}} & \multicolumn{5}{c|}{\textbf{Train Size}} \\ \cline{2-6} 
\multicolumn{1}{|c|}{} & \multicolumn{1}{c}{200} & \multicolumn{1}{c}{600} & \multicolumn{1}{c}{1200} & \multicolumn{1}{c}{2000} & 3000 \\ \hline
Support Vector Clf. & 57.4\% & 63.4\% & 66.3\% & 68.5\% & 69.6\% \\
K-Neighbors & 47.9\% & 52.3\% & 55.9\% & 58.3\% & 60.3\% \\
Depth 8 Dec. Tree & 52.1\% & 57.9\% & 61.9\% & 65.5\% & 66.5\% \\
Random Forest & \textbf{59.0\%} & \textbf{64.0\%} & 65.3\% & 66.5\% & 66.8\% \\
AdaBoost & 48.1\% & 50.1\% & 48.9\% & 49.8\% & 50.4\% \\
Gaussian Pr. Clf. & 47.9\% & 62.5\% & \textbf{66.7\%} & \textbf{69.5\%} & \textbf{72.0\%} \\
3-layer Neural Net & 54.9\% & 61.6\% & 63.9\% & 66.8\% & 68.7\% \\
6-layer Neural Net & 55.3\% & 59.7\% & 62.7\% & 65.5\% & 68.7\% \\ \hline
\end{tabular}%
}
\label{tab:classification}
\end{table}

\paragraph{Comparison of classification using image and parametric data:} 
Next, we seek to understand how different varieties of data impact an algorithm's ability to understand a design. In particular, we examine the relative value of images and parametric data. Three deep neural networks are all trained for the bicycle classification task, however, the first uses only BIKED's parametric data, the second uses only BIKED's image data, and the third uses both simultaneously. Network architectures, training details, and training plots are included in Appendix B2. In this study, the average and highest classification accuracies and F1 scores over ten instantiations are reported in Figure \ref{tab:tfclassification}. The parametric deep neural network's classification performance is notably better than the previous study's neural networks, primarily due to the custom architecture, which includes dropout and batch normalization layers to correct for overfitting. While the three-layer architecture was found to perform best in the previous study, a five-layer network is found to be preferable once measures to combat overfitting are imposed. 

\begin{table}[]
\caption{Classification accuracy and F1 scores for three custom deep networks. Top and average scores are determined over 10 instantiations.}
\renewcommand{\arraystretch}{1}
\centering
\label{tab:tfclassification}
\resizebox{\linewidth}{!}{%
\begin{tabular}{|l|ll|ll|c|}
\hline
\multicolumn{1}{|c|}{\multirow{2}{*}{\textbf{Model}}} & \multicolumn{2}{c|}{\textbf{Average}} & \multicolumn{2}{c|}{\textbf{Top}} & \multirow{2}{*}{\textbf{\begin{tabular}[c]{@{}c@{}}Train\\ Time (s)\end{tabular}}} \\ \cline{2-5}
\multicolumn{1}{|c|}{} & \multicolumn{1}{c}{Acc.} & \multicolumn{1}{c|}{F1} & \multicolumn{1}{c}{Acc.} & \multicolumn{1}{c|}{F1} &  \\ \hline
Parametric DNN & \textbf{72.0\%} & \textbf{48.0\%} & \textbf{73.1\%} & \textbf{51.6\%} & 149.0 \\
Image CNN & 65.6\% & 44.6\% & 66.8\% & 47.9\% & 96.0 \\
Combination NN & 71.0\% & 47.8\% & 72.5\% & 50.0\% & 30.6 \\ \hline
\end{tabular}%
}
\end{table}

The classification accuracy using only parametric data is superior to the accuracy using only images, suggesting that the parametric data is richer in meaningful information than the image data. This is expected since the images are deterministically generated from the parametric data so should have strictly less information. In other words, different parametric vectors may map to the same image, but different images will never map to the same parametric vector. Though we would expect the classifier using both images and parametric data to have comparable performance, it ended up falling slightly short of the parametric classifier due to overfitting, which can be observed in the training plots included in Appendix B2. This overfitting may be explained by the drastically higher input dimensionality of the combined data over the pure parametric data.

\begin{figure}[h]
    \centering
    \includegraphics[scale=.35]{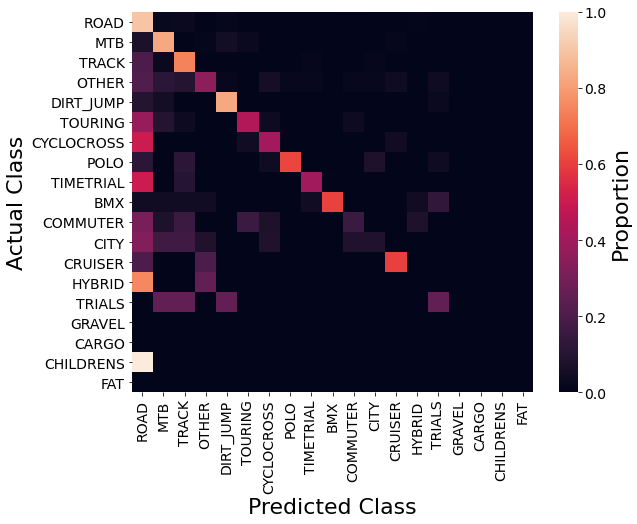}
    \caption{Confusion matrix of parametric DNN predictions on a previously unseen test set. Actual classes are shown on the vertical axis and predictions are shown on the horizontal axis. Classes are sorted from most prevalent (top left) to least prevalent (lower right).}
    \label{fig:Confusion1c}
\end{figure}

\paragraph{Analysis of Classification Results with Confusion Matrices:}
Of the classifiers tested, none attained classification accuracy higher than 75\% or F1 higher than 55\%. For this reason, claiming that machines can truly understand the design of bicycles requires further justification. The formulated problem of bicycle classification is challenging because of the previously mentioned overlapping class labels and imbalance of classes. While aggregate metrics are handy for benchmarking algorithms, examining their exact behavior often requires a more detailed performance analysis. Confusion matrices are often used to report and visualize classification performance. We include confusion matrices that show the proportion of test set designs of a particular class (vertical axis) that are classified into a particular classification (horizontal axis). A perfect classifier's confusion matrix is the identity matrix. Shown in Figure \ref{fig:Confusion1c} is the confusion matrix for the highest performing deep neural network trained on parametric data from the previous subsection. Confusion matrices for the deep image classifier and hybrid classifier are included in Appendix B3.

Examining the confusion matrix, we see that the classifier indeed seems to be struggling due to the overlap and poor definition of some class labels as well as significant class imbalance. Classes with very few designs (fat, children's, cargo, and gravel with 1, 10, 14, and 19 designs respectively) do not have enough training data to be classified well. In addition, three of these classes do not have a single representative model in the test set.  This phenomenon significantly impacts F1 scores, which are calculated with macro averaging (see Appendix B1) and primarily explains the poor F1 performance. Subclasses of road bikes, like `gravel', `timetrial', `cyclocross', and `touring' are often categorized into the general `road' class. Additionally, ambiguous classes like `hybrid', `other', and arguably `city' and `commuter' pose challenges for the classifier since a variety of bicycle styles may fit into these classes. This indicates that the classifier is struggling due to the overlap and poor definition of class labels. From these observations, we can conclude that the classifier's moderate performance is more an indication of the challenges brought about by the construction of the classification problem, rather than a difficulty to understand the design of the bicycle. 

\paragraph{Interpretability analysis to identify design attributes:} 
While we have demonstrated that neural networks can understand the design of bikes and predict class fairly well, they still act as black-box function approximators. Understanding the classification impact of individual parametric features can help designers appreciate the particular factors that define a specific class of bike. For example, one may question what specific features cause the network to assign a particular model to the road bike class instead of the mountain bike class. To study such questions, the final part of this classification study examines the results of a Shapley Additive Explanations (SHAP) analysis ~\cite{NIPS2017_7062}. SHAP locally approximates model predictions over individual input values, effectively capturing the predictive significance of each input parameter. This analysis shows the average impact of a particular feature on the classification probability for each class type.

\begin{figure*}[h]
    \centering
    \includegraphics[scale=.7]{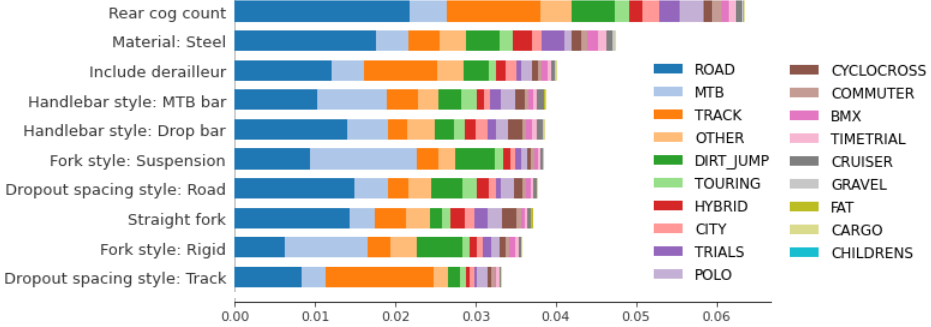}
    \caption{Shapley Additive Explanations analysis of feature significance on design classification result. The top 10 most significant features are shown on the vertical axis and the mean impact to classification probability is shown on the horizontal axis. Since parameters are one-hot encoded, we list boolean values describing whether or not a categorical parameter falls into a particular class as ``Parameter: Class.'' Note that a high significance does not imply a positive correlation.}
    \label{fig:DNNSHAP}
\end{figure*}

Figure~\ref{fig:DNNSHAP} shows the results of the SHAP analysis on the highest performing deep neural network which scores $73.1 \%$ classification accuracy. We observe that each of the most significant parameters besides the number of cogs is a boolean type. Many directly relate to the style of a particular component. The significance of many of these parameters makes intuitive sense. Handlebar, fork, and dropout style are major distinguishing factors between bike classes and most designers of a particular class of bike will choose to use corresponding classes of components for their design. Humans would probably not immediately consider the number of cogs, material, presence of derailleur, or curvature of the fork when classifying bikes, but would likely acknowledge their significance. We can also observe that a few features have a particularly large impact on the probability that a bike is classified as a particular class, like track style dropout spacing for track bikes or suspension forks for mountain bikes. Such SHAP Analyses are reasonably consistent between different instances of the same network architecture, but the importance ranks of the top 20 parameters often fluctuate in ranking by 3-5 places. 

\subsection{Bicycle Synthesis using Variational Autoencoders}\label{synthesis}
In this section, we consider the question of how algorithms can generate previously unseen bicycle designs. While generative methods have been used extensively on images, BIKED's parametric data opens possibilities to directly synthesize designs in the parameter space. We aim to contrast the two methods by considering a variety of bicycle synthesis methods including random generation, interpolation, and extrapolation using Variational Autoencoders (VAEs)~\cite{kingma2013auto, kingma2019introduction}. VAEs encode input data samples into a low dimensional latent space and subsequently decode samples from latent space vectors~\cite{kingma2013auto, kingma2019introduction}. The VAEs can be used for design synthesis by generating new latent space vectors, either through random sampling, interpolation, or extrapolation, then decoding the generated latent space vectors to synthesize new designs. We train two VAEs, one on images and one on parametric data. Since parametric design synthesis results are difficult to qualitatively evaluate, images are generated from synthesized parametric data using the method discussed in Section~\ref{Imregen}. Appendix C1 explains the architecture and training details of the VAEs in detail. 

\begin{figure}
\captionsetup[subfigure]{justification=centering}
    \centering
    \begin{subfigure}[b]{0.15\textwidth}    
        \centering 
        \includegraphics[width=\textwidth]{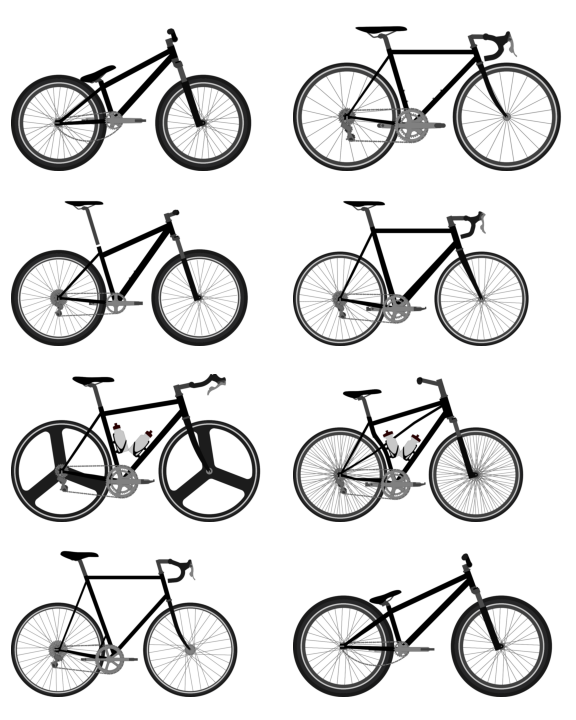}
        \caption[Wheels]%
        {{\small Images of Original Designs}}    
        \label{fig:orig}
    \end{subfigure}
    \hfill
    \begin{subfigure}[b]{0.15\textwidth} 
        \centering
        \includegraphics[width=\textwidth]{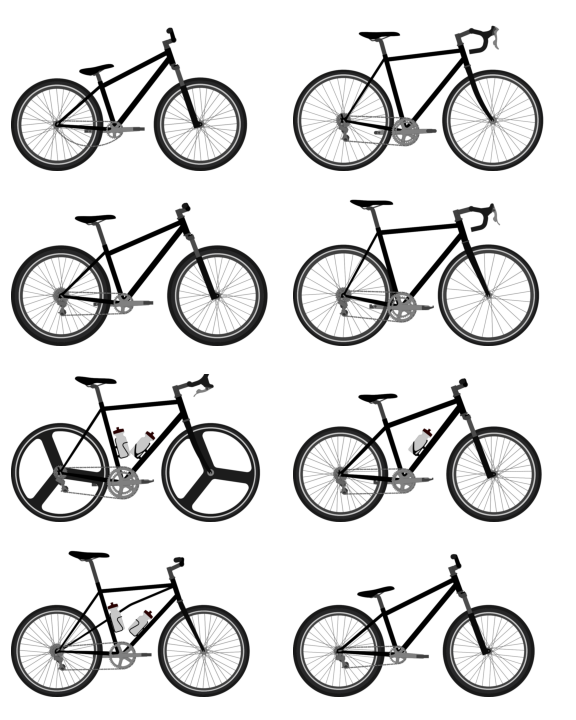}
        \caption[segbike]%
        {{\small Parametric VAE Reconstruction}}    
        \label{fig:dreco}
    \end{subfigure}
    \hfill
    \begin{subfigure}[b]{0.15\textwidth} 
        \centering 
        \includegraphics[width=\textwidth]{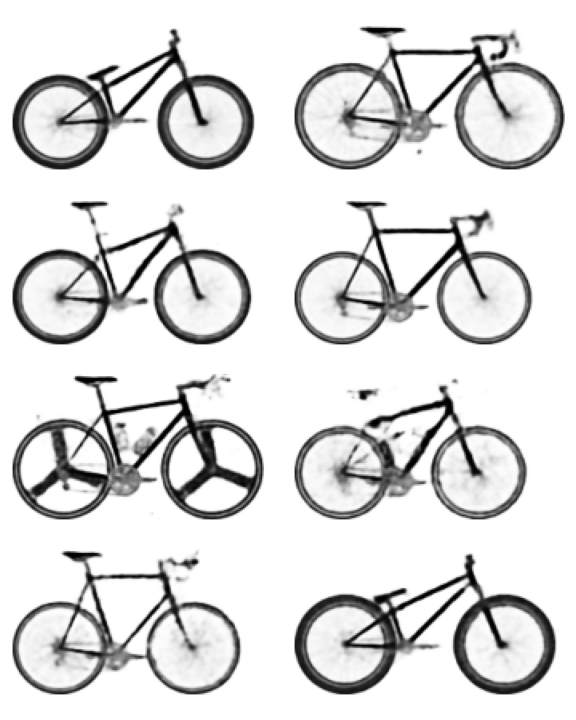}
        \caption[Wheels]%
        {{\small Image VAE Reconstruction}}    
        \label{fig:ireco}
    \end{subfigure}
    \caption[Components]
    {\small Original bike images and images reconstructed by the two Variational Autoencoders.} 
    \label{fig:reconstruction}
\end{figure}

\paragraph{Evaluating trained VAEs using reconstruction quality:}
Examining a VAE's reconstructive ability by encoding and subsequently decoding samples is a common way to evaluate its performance. Figure~\ref{fig:reconstruction} shows the trained VAEs' performance on a reconstruction task on previously unseen data. Compared to the slightly blurry images generated by the image VAE, the parametric VAE tends to generate more realistic images, largely due to the image generation process in BikeCAD. On the other hand, the image VAE tends to reconstruct the input more accurately. This is largely because the image VAE is directly trained to reconstruct visual appearance while the parametric VAE is only trained to do so indirectly. Critical to this observation is the fact that the parametric VAE knows no weighting of the relative importance (or particularly visual importance) of features. For example, the number of teeth on the third rear cog is just as important to the parametric VAE as the handlebar type. Weighting for this relative importance could be a future line of inquiry. Finally, we note that both VAEs struggle to reconstruct bikes with very unconventional features like the inverted handlebar and extra tube of the third bike on the right. This may be a limitation of the quantity of data---when an unconventional feature is present in extremely few designs, the VAEs do not have enough training data to learn to reconstruct these features, especially if these features do not appear in the training set. It may also be a limitation of the VAE model itself, since we experience that the parametric VAE sometimes struggles with training issues like posterior collapse, which can be sensitive to initialization and reduce reliability.

\begin{figure}
\captionsetup[subfigure]{justification=centering}
    \centering
    \begin{subfigure}[b]{0.22\textwidth} 
        \centering
        \includegraphics[width=\textwidth]{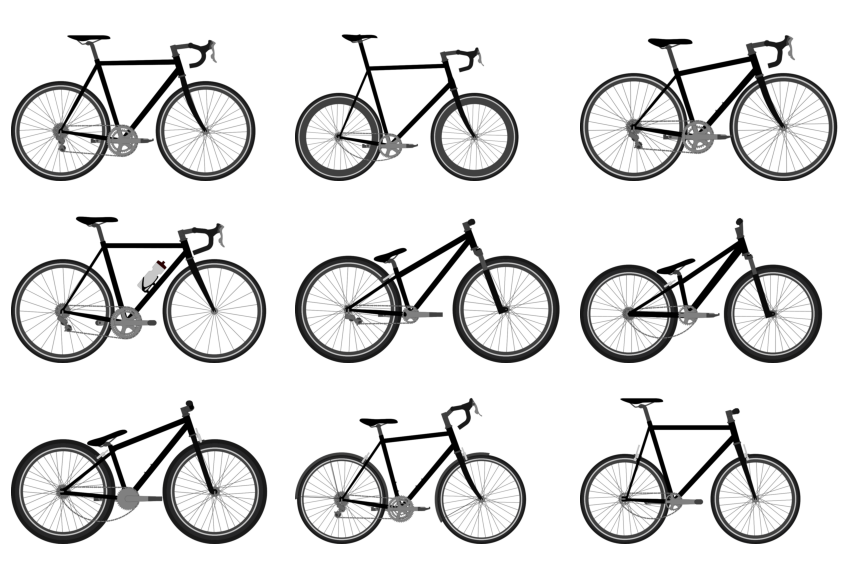}
        {{\small Random Sampling of\\ Original Designs}}    
        \label{fig:origgen}
    \end{subfigure}
    \hfill
    \begin{subfigure}[b]{0.22\textwidth}    
        \centering 
        \includegraphics[width=\textwidth]{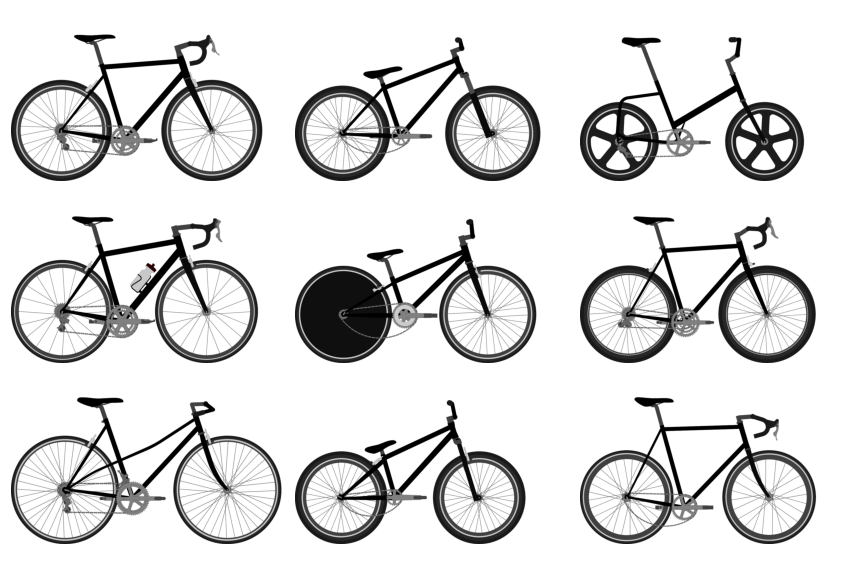}
        {{\small Original Parameter\\Space Interpolation}}    
        \label{fig:interp}
    \end{subfigure}
    \vskip\baselineskip
    \begin{subfigure}[b]{0.22\textwidth} 
        \centering
        \includegraphics[width=\textwidth]{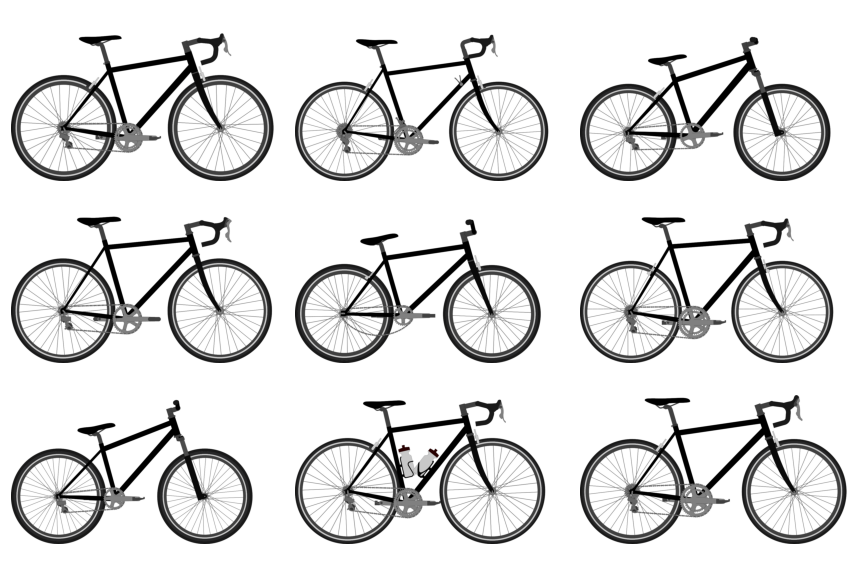}
        {{\small Parametric VAE\\Random Sampling}}    
        \label{fig:vaerand}
    \end{subfigure}
    \hfill
    \begin{subfigure}[b]{0.22\textwidth}    
        \centering 
        \includegraphics[width=\textwidth]{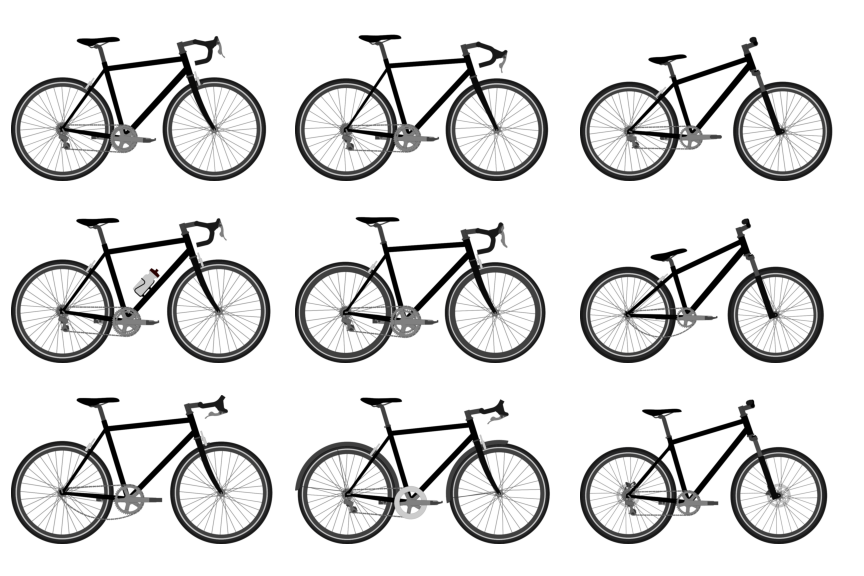}
        {{\small Parametric VAE Latent\\Space Interpolation}}    
        \label{fig:vaeinterpolgen}
    \end{subfigure}
    \vskip\baselineskip
    \begin{subfigure}[b]{0.22\textwidth} 
        \centering
        \includegraphics[width=\textwidth]{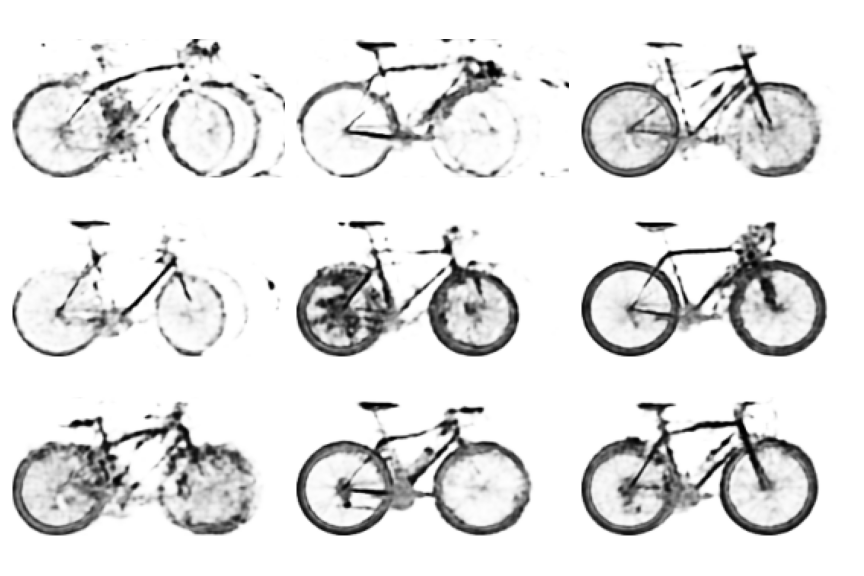}
        {{\small Image VAE\\Random Sampling}}    
        \label{fig:imvaegen}
    \end{subfigure}
    \hfill
    \begin{subfigure}[b]{0.22\textwidth}    
        \centering 
        \includegraphics[width=\textwidth]{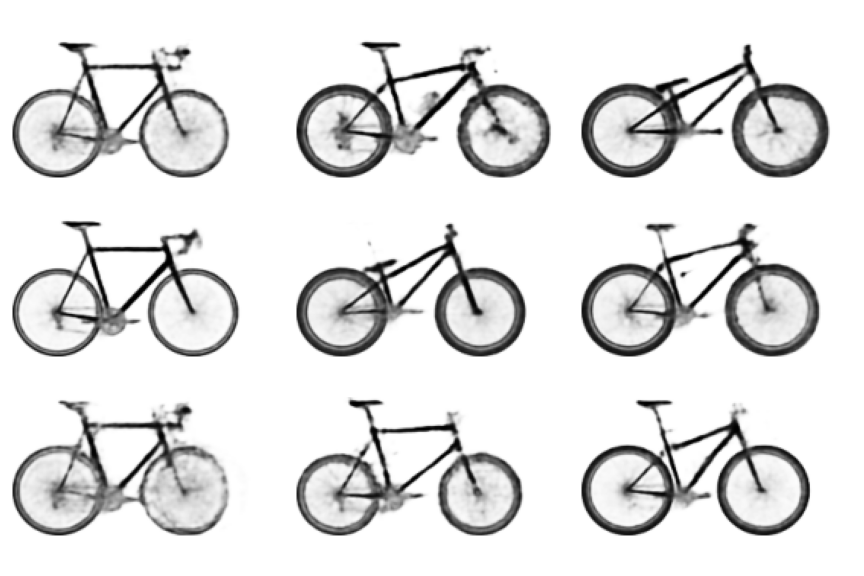}
        {{\small Image VAE Latent\\Space Interpolation}}    
        \label{fig:imvaeinterpgen}
    \end{subfigure}
    \caption[Gen]
    {\small Bike Images randomly generated through random latent space sampling, random latent space interpolation, and random parameter space interpolation.} 
    \label{fig:gen}
\end{figure}

\paragraph{Bicycle synthesis through interpolation and random generation:}
Next, we consider the task of bicycle synthesis. Full design synthesis is a problem with many possible considerations and objectives, but as a starting point, we attempt a qualitative comparison of randomly generated designs from our machine learning models. We test two synthesis methods on each of the trained VAEs. The first of these is randomly sampling a vector from the latent space of the VAE and decoding this vector using the decoder. The random selection is performed by generating a probability distribution function for each value of the encoder's output based on the test data, then sampling from each distribution independently to generate a random latent vector. An alternate approach is to interpolate some arbitrary fraction of the distance between latent vectors encoded from existing designs in the dataset instead of sampling randomly. Finally, we also consider interpolation between dataset designs in the original parameter space. Several designs synthesized using each of these methods are shown in Figure \ref{fig:gen}. Though quantitative evaluation of synthesized designs is not considered in this work, future work may leverage quantitative evaluation metrics that have been used to evaluate Generative Adversarial Networks~\cite{chen2021padgan, xu2019modeling}. We discuss other next steps in generative methods for bicycle design later in this section.

\paragraph{Comparison of interpolation and extrapolation in original parameter space and VAE latent spaces:} 
Next, we dive deeper into interpolation options and further consider extrapolation for the three methods considered above (original parameter space, parametric VAE latent space, and image VAE latent space). We manually select a particularly demonstrative example to showcase, interpolating between a dirt jump and road bike design and extrapolating from the road bike away from the dirt jump bike\footnote{Interpolation/Extrapolation formulation included in Appendix C2}. Results are shown in Figure~\ref{fig:InterExtra}. In the first two columns, we note the discontinuities in the interpolation of categorical components which, due to the nature of the image regeneration process, abruptly switch styles as soon as one style's probability supersedes another. Interpolation between two designs in the parameter space typically works well, but extrapolation quickly becomes unrealistic, generating bikes that are sometimes disconnected, as shown in the bottom two images on the left. Interpolation and extrapolation both work well in the latent space of the parametric VAE, with extrapolation results generally being quite realistic. In the example shown, extrapolating on the road bike in a direction opposite the dirt jump bike shifts the design to begin resembling a timetrial (triathlon) bike. However, the limited reconstructive capability of the parametric VAE detracts from the fidelity of the newly generated designs with respect to the original designs. Extrapolation based on the image VAE is very poor and is further exacerbated by the limited fidelity of the image VAE's reconstruction, which is particularly poor in this example. Interpolation in the image VAE's latent space has other undesirable properties, such as the visual overlap of the source designs rather than a gradual change in geometry. Based on these qualitative results, interpolation based on a simple image VAE does not seem to be an effective method for quality bicycle synthesis. 
\paragraph{Further advancement of Generative Methods for Bicycles}
We envision many opportunities for further development of generative methods for bicycles. Other generative methods like Generative Adversarial Networks (GANs)~\cite{goodfellow2014generative} may be promising alternatives to Variational Autoencoders (VAEs). BIKED's component labels and images also open possibilities for hierarchical design methods or approaches based on symbolic reasoning. Conditional approaches based on bike class may also be viable. We also note that reliable synthesis of designs will likely require rule-based feasibility checking of geometric validity and manufacturing constraints in a sort of ``expert-system-based'' approach. In addition to feasibility checking, performance evaluation of considerations like structural stiffness, aerodynamics, weight, and ergonomics will be important in any performance-aware synthesis method and will be an interesting avenue for future research. Incorporating BIKED-based automated feasibility checking, performance evaluation, and optimization methods with advanced generative methods is a promising avenue towards the ultimate goal of full performance-aware bicycle synthesis. 

\begin{figure}[h]
    \centering
    \includegraphics[scale=.44]{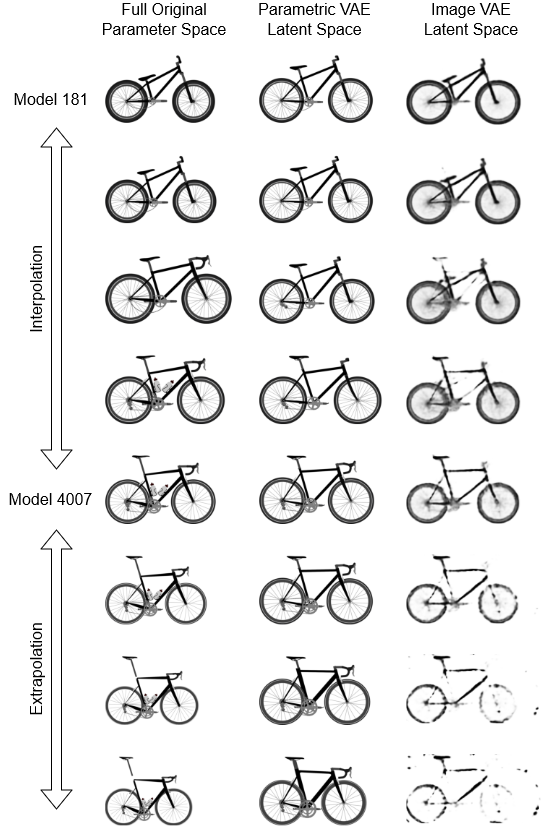}
    \caption{Interpolation and extrapolation results between a road bike and dirt jump bike. The left side of the figure shows interpolation and extrapolation results carried out in the original parameter space between original designs. The center and right show results from interpolation and extrapolation carried out between latent space representations of the original designs encoded by the two Variational Autoencoders, then decoded and reprocessed (when applicable) into images.}
    \label{fig:InterExtra}
\end{figure}

\section{Strengths and Novelty}
Compared to existing 3D model and topology optimization datasets that are commonly used in the data-driven design field, BIKED emphasizes comprehensive design information. This makes BIKED well suited to a multitude of data-driven design tasks that would not be feasible using other datasets. BIKED's detailed component images and parametric data may furthermore support hierarchical methods that would not be supported by other datasets.  

BIKED's designs are all hand-designed and, as discussed in Section~\ref{bikecad}, the inbuilt software features help ensure that the majority of BIKED designs are of good quality. BIKED is extensively processed, organized, and documented to make BIKED as user-friendly as possible.

\section{Limitations}
As is the case in most datasets, BIKED has several sources of bias to note. The fact that all designs were modeled in BikeCAD is a potential source of bias, since BikeCAD's design tools and parameterization scheme might make certain designs more or less easy to model. Furthermore, there is a bias towards the default bike designs that often serve as the starting point in a user's design process. This bias may be noticeable at a high level in overall design characteristics, or at a low level with users leaving certain dimensions at their default values. This is especially true for more niche parameters that have little to no visual impact on the appearance of the bike model, where novice users may not even understand the parameter's meaning. This bias causes many parameters to have uniform values across the vast majority of designs, which can then cause issues, such as posterior collapse while training Variational Autoencoders. Additionally, BIKED has a temporal bias which is demonstrated and discussed in more detail in Appendix D1. Since model numbers largely correlate to the BikeCAD version originally used to create the design, updates to the software that introduced new parameters or updated old default values are likely the root cause. 

We also reiterate some of the potential biases discussed in Section~\ref{paramredux}. Dropping bikes with particular features like tandem components or extra nonstandard tubes may have introduced biases. Additionally, manually overriding component visibility of critical design components could have unintentionally included components that were not intended to be part of the designer's vision. We have worked closely with bicycle experts to use domain knowledge to alleviate as many sources of bias as possible and plan to continue refining our methods and dataset to provide a rich and accurate resource to inspire research on data-driven bicycle design. 

We also note that, although BikeCAD is used by many experienced designers, there is no guarantee of quality or performance of any bike design. Some designs are created to be facetious or highly unrealistic. We would also advise users that the raw unprocessed dataset may contain offensive or proprietary imagery and language since designers are free to use terminology and graphical designs of their choice. 
\section{Future Research Directions}\label{future}

Aside from the three key research questions discussed, we envision many exciting ways to harness BIKED in data-driven design research and bicycling science beyond the initial research presented in this paper. Researchers studying bicycle performance can leverage BIKED's parametric data as an input to multi-physics simulations or develop data-driven regression models. Surrogate models of bicycle frame stiffness or aerodynamic drag could accelerate an iterative design process by eliminating computational bottlenecks caused by Finite Element Analysis (FEA) or Computational Fluid Dynamics (CFD) simulation. Efficient surrogate models can enable high-level optimization tasks like Multidisciplinary Design Optimization (MDO). Rapid performance evaluation methods would be an integral component of any comprehensive performance-aware bicycle synthesis process. BIKED's diversity of bicycle designs also lends itself to studies of customer preference and modeling of this data in conjunction with bicycle classes and parametric properties. The fact that BIKED designs were created largely by bike enthusiasts and tradespeople may be of intrinsic use as a study of customer preference. A more advanced study may expand on BIKED's models with further customer preference surveys for a more detailed study of customer preference in the bicycle market. 

\section{Conclusion}
In this paper, we have presented BIKED, a dataset for data-driven bicycle design. BIKED features 4500 bicycle designs providing full bicycle assembly images, segmented component images, and extensive parametric data to enable various data-driven design approaches. Throughout the paper, we discussed the various processing steps taken to curate the data, and discussed key data-driven applications that we expect BIKED to facilitate. We demonstrated some initial exploration in these directions while simultaneously uncovering features of the dataset and providing baseline performance for common machine learning methods. In the first of these studies, we considered unsupervised dimensionality reduction methods to gain various insights about the dataset. In the second, we trained numerous classification models to predict bicycle class, including 10 baseline models using parametric data and 3 tuned deep neural networks to contrast classification performance using different types of input data. Using one of these classification models, we performed a Shapley Additive Explanations analysis to understand the impact of individual parameters on classification predictions. Next, we trained two Variational Autoencoders to perform a variety of different tasks on both parametric and image data, including design reconstruction, latent space random sampling, interpolation, and extrapolation. Finally, we discussed key strengths and novelty as well as potential limitations and biases in the dataset. 

BIKED goes above and beyond existing datasets, providing extensive design information of various types and promising to enable a slew of novel data-driven design applications. In our discussion, we documented the process to curate the dataset, explored BIKED's features, demonstrated baseline performance for common algorithms, and attempted to inspire researchers with sample applications. We hope that BIKED will prove a valuable resource to researchers in the data-driven design community and will support novel and innovative approaches in this rapidly growing field. 

\section{Acknowledgments}
We would like to thank Professor Daniel Frey for his input and guidance throughout the project. We would like to thank Kris Vu for assisting with sourcing files from the BikeCAD archive and Amin Heyrani Nobari for assisting with the exporting of component images. Finally, we would like to acknowledge MathWorks for supporting this research. 
\bibliographystyle{asmems4}
\bibliography{bibliography}

\newpage
\appendix       
\section{Appendix A1: Dataset Details}
We include a table listing a few sample parameters from each type in Table \ref{tab:params}. We note that any of these can be used as the labels in regression or classification tasks, depending on the researcher's goals, though certain parameters like ``bike style'' may make intuitive sense for many applications. 

\begin{table*}[]
\caption{Example parameters of each of the four datatype present in BIKED.}
\label{tab:params}
\begin{tabular}{|l|l|l|l|}  \hline
\textbf{Continuous} & \textbf{Discrete} & \textbf{Categorical} & \textbf{Boolean} \\ \hline
Top tube length & Number of rear cogs & Wheel type & Include racks (Y/N) \\ \hline
Bottom bracket drop distance & Number of teeth on first rear cog & Handlebar style & Symmetric chain stays (Y/N) \\ \hline
Saddle height & Number of spokes on front wheel & Bicycle style & Include fenders (Y/N) \\ \hline
\end{tabular}
\end{table*}

\section*{Appendix B1: Training Details and extended results of Simple Classifiers}
We include a more detailed study of classification performance on BIKED's parametric data for a baseline set of 12 classifiers with 5 sizes of training data. For each classifier and train set size, we record the average and top accuracy, recall, precision, and F1, as well as the average evaluation time over 10 instantiations. Each classification model is trained on 200, 600, 1200, 2000, and 3000 randomly sampled designs, leaving the remainder of the 4512 processed designs for testing. We do not include a validation set since no tuning is performed. Only parametric data is used in this study. Each of the 10 instantiations consists of a new model initialization and a different random train-test split. Precision, recall, and F1 are calculated using macro averaging, i.e. calculating recall, precision, and F1 for each individual class and subsequently performing an unweighted average over the classes. If no representative models of a class appear in the test data, we set precision, recall, and F1 to 0 for the class. Table~\ref{tab:bigclass} contains the results of this study. The training was performed on a computer with Xeon Silver 4215R CPU, 64 Gb RAM, and two NVIDIA Quadro RTX4000 GPUs. Key parameters of classification models are listed below. For further implementation details of the methods tested, we refer the reader to our provided code. 
\begin{enumerate}
    \item Neural networks tested have 200 neurons per layer, use Rectified Linear Unit (ReLu) activation functions, and are trained using the adam optimizer~\cite{kingma2014adam}. 
    \item K-Neighbors-Classifier uses 5 neighbors
    \item Random Forest Classifier uses 50 estimators and a maximum depth of 7
    \item Gaussian Process Classifier uses the Radial-basis function kernel with a length parameter of 1 and scaling factor of 1
    \item SVC classifier uses a linear kernel
\end{enumerate}

\begin{table*}[]
\renewcommand{\arraystretch}{1.0}
\centering
\caption{Classification accuracy, precision, recall, and F1 for 12 classifiers using BIKED's parametric data.}
\label{tab:bigclass}
\resizebox{\textwidth}{!}{%
\begin{tabular}{|
>{\columncolor[HTML]{FFFFFF}}c |
>{\columncolor[HTML]{FFFFFF}}l |
>{\columncolor[HTML]{FFFFFF}}c 
>{\columncolor[HTML]{FFFFFF}}c 
>{\columncolor[HTML]{FFFFFF}}c 
>{\columncolor[HTML]{FFFFFF}}c |
>{\columncolor[HTML]{FFFFFF}}c 
>{\columncolor[HTML]{FFFFFF}}c 
>{\columncolor[HTML]{FFFFFF}}c 
>{\columncolor[HTML]{FFFFFF}}c |
>{\columncolor[HTML]{FFFFFF}}c |}
\hline
\cellcolor[HTML]{FFFFFF} & \multicolumn{1}{c|}{\cellcolor[HTML]{FFFFFF}} & \multicolumn{4}{c|}{\cellcolor[HTML]{FFFFFF}\textbf{Average}} & \multicolumn{4}{c|}{\cellcolor[HTML]{FFFFFF}\textbf{Top}} & \cellcolor[HTML]{FFFFFF} \\ \cline{3-10}
\multirow{-2}{*}{\cellcolor[HTML]{FFFFFF}\textbf{Train Size}} & \multicolumn{1}{c|}{\multirow{-2}{*}{\cellcolor[HTML]{FFFFFF}\textbf{Model}}} & Accuracy & Recall & Precision & F1 & Accuracy & Recall & Precision & F1 & \multirow{-2}{*}{\cellcolor[HTML]{FFFFFF}\begin{tabular}[c]{@{}c@{}} \textbf{Average} \\\textbf{Train Time (s)} \end{tabular}} \\ \hline
\cellcolor[HTML]{FFFFFF} & Support Vector Clf. & 57.4\% & 16.3\% & 33.8\% & 17.0\% & 59.9\% & 18.3\% & 41.3\% & 20.2\% & 9.3 \\
\cellcolor[HTML]{FFFFFF} & Naïve Bayes & 43.7\% & 9.3\% & 30.7\% & 9.8\% & 46.9\% & 11.5\% & 41.0\% & 12.3\% & 4.3 \\
\cellcolor[HTML]{FFFFFF} & K-Neighbors & 47.9\% & 13.3\% & 19.4\% & 12.8\% & 51.4\% & 17.1\% & 23.4\% & 15.2\% & 11.8 \\
\cellcolor[HTML]{FFFFFF} & Depth 5 Dec. Tree & 52.0\% & 19.3\% & 22.4\% & 19.2\% & 55.7\% & 23.1\% & 30.4\% & 22.1\% & \textbf{0.2} \\
\cellcolor[HTML]{FFFFFF} & Depth 8 Dec. Tree & 52.1\% & 20.9\% & 22.4\% & 20.7\% & 57.4\% & \textbf{28.0\%} & 27.2\% & \textbf{26.5\%} & \textbf{0.2} \\
\cellcolor[HTML]{FFFFFF} & Random Forest & \textbf{59.0\%} & 18.6\% & \textbf{43.0\%} & 19.6\% & \textbf{61.8\%} & 21.0\% & \textbf{48.9\%} & 21.6\% & 0.4 \\
\cellcolor[HTML]{FFFFFF} & AdaBoost & 48.1\% & 10.4\% & 10.7\% & 7.8\% & 51.1\% & 13.0\% & 18.3\% & 9.9\% & 4.7 \\
\cellcolor[HTML]{FFFFFF} & Gaussian Pr. Clf. & 47.9\% & 9.9\% & 8.4\% & 7.9\% & 58.8\% & 23.5\% & 29.0\% & 21.4\% & 122.8 \\
\cellcolor[HTML]{FFFFFF} & 3-layer Neural Net & 54.9\% & 19.8\% & 32.7\% & 21.0\% & 56.2\% & 21.6\% & 40.0\% & 22.5\% & 4.2 \\
\cellcolor[HTML]{FFFFFF} & 4-layer Neural Net & 54.8\% & 19.5\% & 30.0\% & 20.6\% & 56.9\% & 22.5\% & 37.6\% & 23.1\% & 2.7 \\
\cellcolor[HTML]{FFFFFF} & 5-layer Neural Net & 54.6\% & 20.0\% & 27.5\% & 20.9\% & 56.9\% & 23.4\% & 32.5\% & 24.0\% & 2.5 \\
\multirow{-12}{*}{\cellcolor[HTML]{FFFFFF}200} & 6-layer Neural Net & 55.3\% & \textbf{21.2\%} & 28.7\% & \textbf{22.2\%} & 57.1\% & 25.4\% & 33.2\% & 25.5\% & 2.3 \\ \hline
\cellcolor[HTML]{FFFFFF} & Support Vector Clf. & 63.4\% & 23.4\% & 45.4\% & 25.6\% & 64.7\% & 25.9\% & 54.7\% & 28.7\% & 23.6 \\
\cellcolor[HTML]{FFFFFF} & Naïve Bayes & 43.3\% & 14.7\% & 37.6\% & 16.8\% & 45.0\% & 16.9\% & 44.2\% & 19.0\% & 4.3 \\
\cellcolor[HTML]{FFFFFF} & K-Neighbors & 52.3\% & 18.1\% & 27.9\% & 19.1\% & 53.9\% & 20.2\% & 34.9\% & 21.8\% & 34.3 \\
\cellcolor[HTML]{FFFFFF} & Depth 5 Dec. Tree & 59.0\% & 24.1\% & 29.1\% & 23.9\% & 62.3\% & 28.5\% & 37.0\% & 27.6\% & \textbf{0.2} \\
\cellcolor[HTML]{FFFFFF} & Depth 8 Dec. Tree & 57.9\% & 26.9\% & 30.2\% & 27.3\% & 59.6\% & 29.3\% & 36.4\% & 31.4\% & 0.3 \\
\cellcolor[HTML]{FFFFFF} & Random Forest & \textbf{64.0\%} & 23.6\% & \textbf{50.3\%} & 25.5\% & \textbf{65.4\%} & 27.1\% & 57.8\% & 28.5\% & 0.4 \\
\cellcolor[HTML]{FFFFFF} & AdaBoost & 50.1\% & 10.8\% & 9.2\% & 8.2\% & 53.9\% & 15.9\% & 13.8\% & 12.3\% & 4.7 \\
\cellcolor[HTML]{FFFFFF} & Gaussian Pr. Clf. & 62.5\% & 25.1\% & 39.9\% & 25.6\% & 65.0\% & 27.6\% & \textbf{61.1\%} & 29.0\% & 454.8 \\
\cellcolor[HTML]{FFFFFF} & 3-layer Neural Net & 61.6\% & 29.0\% & 43.0\% & \textbf{31.6\%} & 62.7\% & 30.9\% & 48.3\% & \textbf{34.5\%} & 14.5 \\
\cellcolor[HTML]{FFFFFF} & 4-layer Neural Net & 60.8\% & \textbf{29.0\%} & 42.1\% & 31.4\% & 61.7\% & \textbf{32.0\%} & 47.7\% & 33.7\% & 10.1 \\
\cellcolor[HTML]{FFFFFF} & 5-layer Neural Net & 58.6\% & 27.4\% & 35.0\% & 28.7\% & 62.4\% & 30.7\% & 43.7\% & 32.9\% & 8.5 \\
\multirow{-12}{*}{\cellcolor[HTML]{FFFFFF}600} & 6-layer Neural Net & 59.7\% & 29.0\% & 38.3\% & 30.7\% & 62.0\% & 31.1\% & 46.6\% & 33.9\% & 8.5 \\ \hline
\cellcolor[HTML]{FFFFFF} & Support Vector Clf. & 66.3\% & 28.5\% & 49.1\% & 31.4\% & \textbf{68.4\%} & 31.9\% & \textbf{60.5\%} & 36.0\% & 40.7 \\
\cellcolor[HTML]{FFFFFF} & Naïve Bayes & 41.8\% & 20.8\% & 32.2\% & 22.2\% & 42.8\% & 22.7\% & 40.8\% & 23.8\% & 3.6 \\
\cellcolor[HTML]{FFFFFF} & K-Neighbors & 55.9\% & 23.5\% & 37.3\% & 25.4\% & 56.7\% & 26.5\% & 43.6\% & 29.3\% & 59.1 \\
\cellcolor[HTML]{FFFFFF} & Depth 5 Dec. Tree & 62.8\% & 26.3\% & 30.0\% & 26.1\% & 64.3\% & 28.6\% & 33.8\% & 29.5\% & \textbf{0.3} \\
\cellcolor[HTML]{FFFFFF} & Depth 8 Dec. Tree & 61.9\% & 32.7\% & 38.3\% & 33.6\% & 63.5\% & 36.1\% & 42.7\% & 36.7\% & \textbf{0.3} \\
\cellcolor[HTML]{FFFFFF} & Random Forest & 65.3\% & 25.2\% & \textbf{51.6\%} & 26.8\% & 67.0\% & 26.3\% & 59.4\% & 28.8\% & 0.4 \\
\cellcolor[HTML]{FFFFFF} & AdaBoost & 48.9\% & 10.5\% & 7.5\% & 7.8\% & 51.1\% & 12.0\% & 9.9\% & 9.1\% & 4 \\
\cellcolor[HTML]{FFFFFF} & Gaussian Pr. Clf. & \textbf{66.7}\% & 32.2\% & 46.4\% & 33.8\% & 67.8\% & 35.5\% & 59.2\% & 37.1\% & 1204.8 \\
\cellcolor[HTML]{FFFFFF} & 3-layer Neural Net & 63.9\% & 34.7\% & 50.4\% & \textbf{38.4\%} & 65.9\% & 37.9\% & 53.0\% & \textbf{41.8\%} & 28.9 \\
\cellcolor[HTML]{FFFFFF} & 4-layer Neural Net & 63.2\% & 34.7\% & 45.8\% & 37.6\% & 64.0\% & 37.5\% & 49.8\% & 40.5\% & 19.7 \\
\cellcolor[HTML]{FFFFFF} & 5-layer Neural Net & 63.3\% & \textbf{34.9\%} & 45.3\% & 37.1\% & 64.9\% & 37.5\% & 53.1\% & 40.1\% & 16.3 \\
\multirow{-12}{*}{\cellcolor[HTML]{FFFFFF}1200} & 6-layer Neural Net & 62.7\% & 34.2\% & 43.1\% & 35.7\% & 65.1\% & \textbf{38.4\%} & 51.2\% & 40.3\% & 12.7 \\ \hline
\cellcolor[HTML]{FFFFFF} & Support Vector Clf. & 68.5\% & 31.9\% & 54.9\% & 35.6\% & 69.9\% & 33.8\% & \textbf{67.0\%} & 38.9\% & 54.1 \\
\cellcolor[HTML]{FFFFFF} & Naïve Bayes & 40.0\% & 24.8\% & 29.6\% & 24.4\% & 41.8\% & 26.4\% & 35.7\% & 26.4\% & 2.2 \\
\cellcolor[HTML]{FFFFFF} & K-Neighbors & 58.3\% & 27.3\% & 43.0\% & 30.2\% & 60.2\% & 29.7\% & 48.4\% & 33.6\% & 78 \\
\cellcolor[HTML]{FFFFFF} & Depth 5 Dec. Tree & 63.7\% & 26.7\% & 32.6\% & 26.7\% & 66.3\% & 30.9\% & 40.2\% & 31.1\% & \textbf{0.3} \\
\cellcolor[HTML]{FFFFFF} & Depth 8 Dec. Tree & 65.5\% & 32.8\% & 41.1\% & 34.8\% & 67.0\% & 35.2\% & 48.8\% & 38.2\% & \textbf{0.3} \\
\cellcolor[HTML]{FFFFFF} & Random Forest & 66.5\% & 26.6\% & 51.3\% & 28.5\% & 67.7\% & 27.3\% & 60.9\% & 29.4\% & 0.4 \\
\cellcolor[HTML]{FFFFFF} & AdaBoost & 49.8\% & 11.0\% & 7.9\% & 8.2\% & 52.9\% & 17.6\% & 13.7\% & 12.9\% & 4.1 \\
\cellcolor[HTML]{FFFFFF} & Gaussian Pr. Clf. & \textbf{69.5\%} & 37.9\% & 54.6\% & 39.7\% & \textbf{71.8\%} & 43.3\% & 62.8\% & 46.2\% & 2686.4 \\
\cellcolor[HTML]{FFFFFF} & 3-layer Neural Net & 66.8\% & \textbf{41.0\%} & \textbf{55.3\%} & \textbf{44.8\%} & 67.7\% & 43.5\% & 64.2\% & 47.8\% & 47 \\
\cellcolor[HTML]{FFFFFF} & 4-layer Neural Net & 65.7\% & 40.1\% & 53.1\% & 43.4\% & 67.6\% & 43.4\% & 62.0\% & \textbf{48.0\%} & 26.1 \\
\cellcolor[HTML]{FFFFFF} & 5-layer Neural Net & 65.4\% & 38.3\% & 50.1\% & 40.6\% & 69.5\% & 42.8\% & 56.7\% & 45.3\% & 20.1 \\
\multirow{-12}{*}{\cellcolor[HTML]{FFFFFF}2000} & 6-layer Neural Net & 65.5\% & 38.8\% & 48.0\% & 40.5\% & 68.4\% & \textbf{46.5\%} & 53.0\% & 46.2\% & 20.1 \\ \hline
\cellcolor[HTML]{FFFFFF} & Support Vector Clf. & 69.6\% & 34.3\% & 55.7\% & 38.0\% & 70.7\% & 37.8\% & 68.3\% & 43.1\% & 58.7 \\
\cellcolor[HTML]{FFFFFF} & Naïve Bayes & 38.3\% & 31.5\% & 28.5\% & 27.9\% & 40.9\% & 36.3\% & 33.1\% & 32.3\% & 1.4 \\
\cellcolor[HTML]{FFFFFF} & K-Neighbors & 60.3\% & 30.8\% & 45.2\% & 32.9\% & 61.4\% & 35.5\% & 52.1\% & 35.7\% & 64.3 \\
\cellcolor[HTML]{FFFFFF} & Depth 5 Dec. Tree & 64.6\% & 29.9\% & 35.4\% & 29.7\% & 65.7\% & 32.6\% & 40.9\% & 31.1\% & \textbf{0.3} \\
\cellcolor[HTML]{FFFFFF} & Depth 8 Dec. Tree & 66.5\% & 34.6\% & 43.6\% & 36.3\% & 67.3\% & 39.0\% & 47.2\% & 40.7\% & 0.4 \\
\cellcolor[HTML]{FFFFFF} & Random Forest & 66.8\% & 27.2\% & 50.5\% & 28.9\% & 69.6\% & 31.5\% & 58.5\% & 33.9\% & 0.5 \\
\cellcolor[HTML]{FFFFFF} & AdaBoost & 50.4\% & 10.6\% & 7.6\% & 7.8\% & 55.0\% & 14.0\% & 15.7\% & 11.4\% & 4.3 \\
\cellcolor[HTML]{FFFFFF} & Gaussian Pr. Clf. & \textbf{72.0\%} & \textbf{42.7\%} & \textbf{63.2\%} & 47.3\% & \textbf{74.4\%} & 47.7\% & \textbf{73.3\%} & 53.0\% & 5250.4 \\
\cellcolor[HTML]{FFFFFF} & 3-layer Neural Net & 68.7\% & 45.6\% & 60.6\% & \textbf{49.9\%} & 71.2\% & 51.2\% & 66.4\% & \textbf{55.0\%} & 50.9 \\
\cellcolor[HTML]{FFFFFF} & 4-layer Neural Net & 67.5\% & 44.1\% & 57.1\% & 47.4\% & 69.5\% & 49.8\% & 68.1\% & 53.5\% & 28.5 \\
\cellcolor[HTML]{FFFFFF} & 5-layer Neural Net & 68.6\% & 44.6\% & 55.0\% & 47.5\% & 69.9\% & \textbf{51.6\%} & 62.2\% & 52.0\% & 31.7 \\
\multirow{-12}{*}{\cellcolor[HTML]{FFFFFF}3000} & 6-layer Neural Net & 68.7\% & 47.2\% & 54.0\% & 48.7\% & 70.9\% & 51.5\% & 59.5\% & 53.2\% & 29.3 \\ \hline
\end{tabular}%
}
\end{table*}

\section*{Appendix B2: Architecture and training details of deep classifiers}
\begin{figure}[h]
    \centering
    \includegraphics[scale=.048]{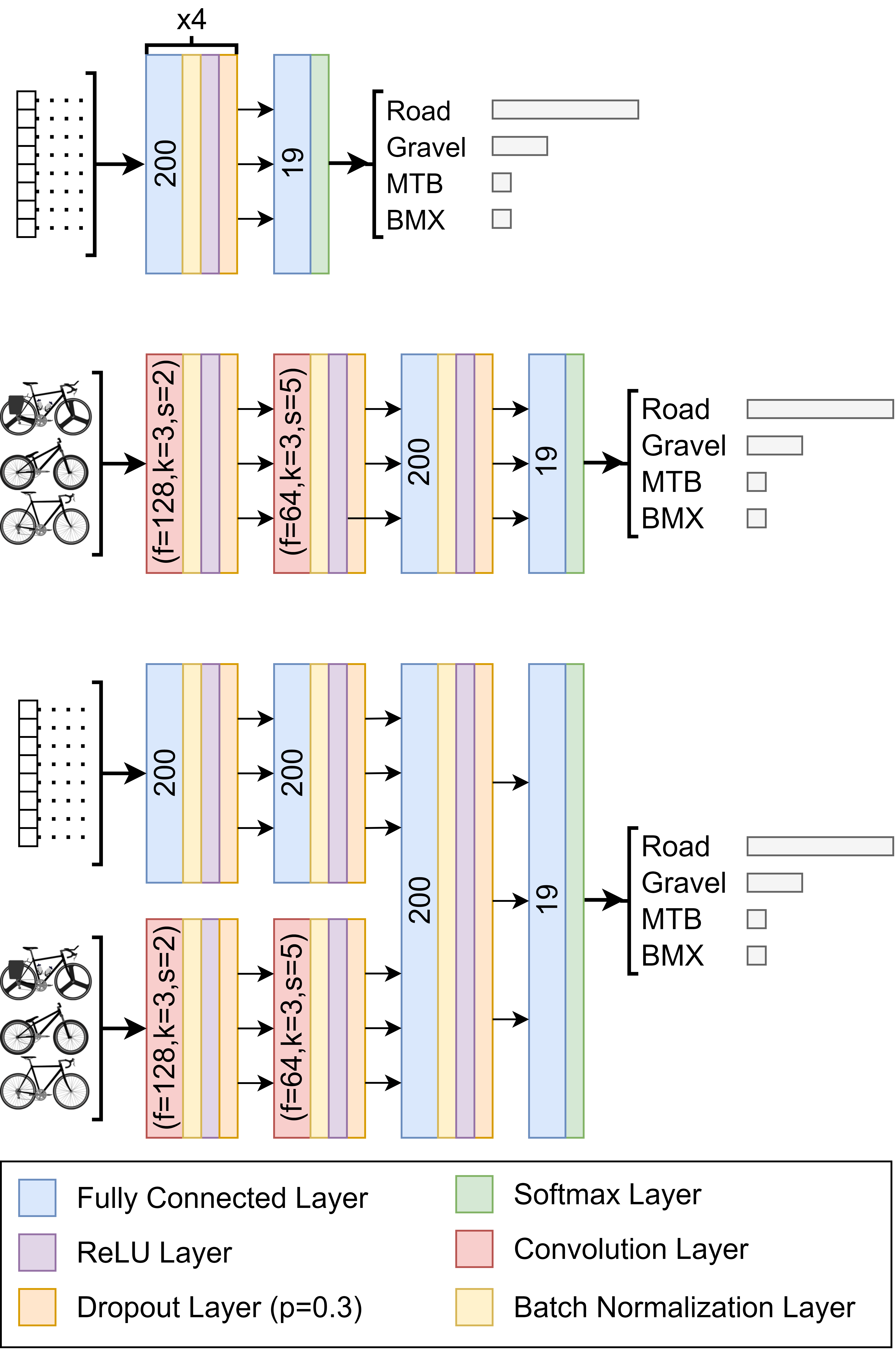}
    \caption{Network diagrams of the three classification networks implemented.}
    \label{fig:NNs}
\end{figure}

\begin{figure}[h]
    \centering
    \includegraphics[scale=.45]{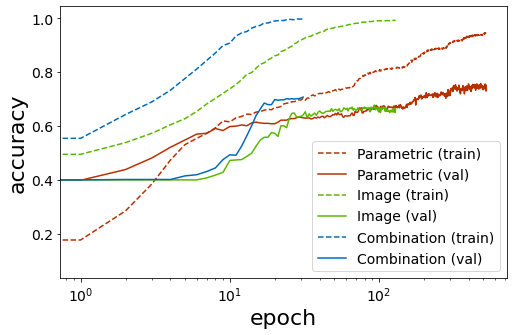}
    \caption{Accuracy vs training epoch (log scaled) for the three tuned classification networks implemented.}
    \label{fig:accuracy}
\end{figure}

\begin{figure}[h]
    \centering
    \includegraphics[scale=.4]{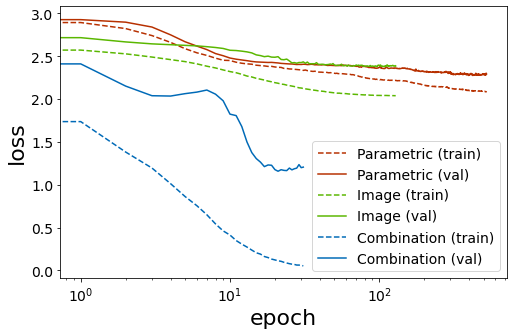}
    \caption{Loss vs training epoch (log scaled) for the three classification networks implemented.}
    \label{fig:loss}
\end{figure}

The architectures of the three custom neural networks are shown in Figure~\ref{fig:NNs}. The number of neurons in every fully connected layer is labeled in the diagram. Similarly, the number of filters (f), kernel size (k), and stride size (s) are shown for every convolution layer. Dropout layers are labeled according to dropout probability. Models were trained using the Adam optimizer~\cite{kingma2014adam} with a learning rate of $10^{-4}$ using a categorical cross-entropy loss. A batch size of 100 was used for all models. The 4512 models were split into training, validation, and test sets with a ratio of 70:15:15 (3158:677:677). Unlike the study in B1, only one train-test-validation set was used for every instantiation of every model. The training was performed on a computer with Ryzen 9 5950x, 32 GB RAM, and Nvidia RTX3080 GPU. 
Training plots of the top-performing instantiations of the networks are included in Figures \ref{fig:accuracy} and \ref{fig:loss}. The training was halted after 100, 20, and 10 epochs without improvement for the parametric, image, and combination classifiers respectively.

\section*{Appendix B3: Classification Confusion Matrix}
\begin{figure}[h]
    \centering
    \includegraphics[scale=.35]{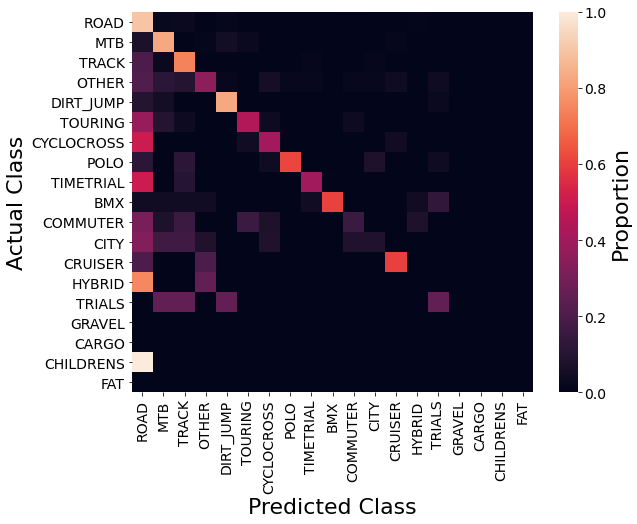}
    \caption{Confusion matrix of Parametric DNN predictions}
    \label{fig:Confusion1}
\end{figure}

\begin{figure}[h]
    \centering
    \includegraphics[scale=.35]{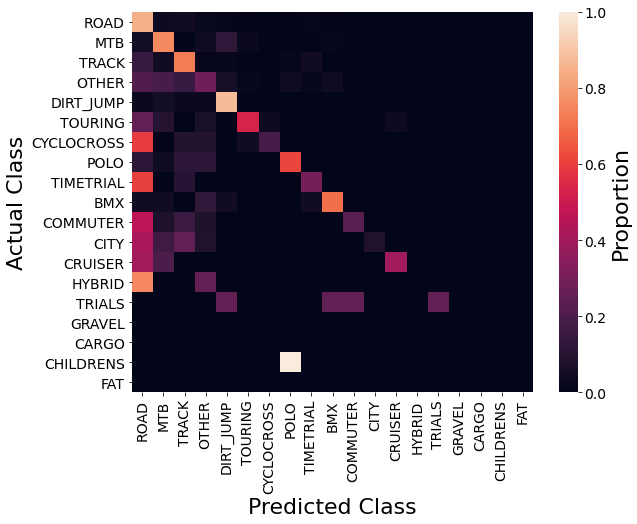}
    \caption{Confusion matrix of Combination NN predictions}
    \label{fig:Confusion3}
\end{figure}

Shown in Figures~\ref{fig:Confusion1} (repeated from the main text) through~\ref{fig:Confusion3} are confusion matrices of the classification performance of the three custom-built deep classifiers. We can appreciate that all three deep classifiers struggle with the class imbalance and class overlap issues discussed in Section \ref{classification}. We also note that classes of bikes that are visually similar to others are more difficult for the image-based CNN to classify. For example, track bikes are primarily identifiable by their single-speed drivetrain, but since the cassette of the bike is not visually prominent, track bikes are more challenging for the CNN compared to the other two classifiers. 
\begin{figure}[h]
    \centering
    \includegraphics[scale=.35]{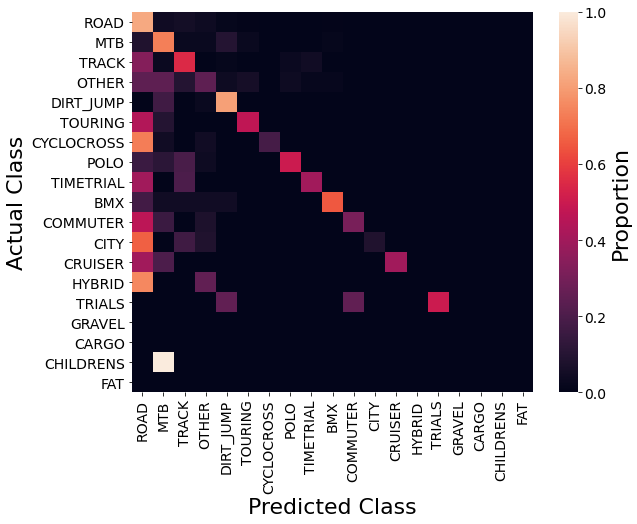}
    \caption{Confusion matrix of Image CNN predictions}
    \label{fig:Confusion2}
\end{figure}

\newpage

\section*{Appendix C1: Architecture and training details of VAEs}\label{appeninterp}

\begin{figure}[h]
    \centering
    \includegraphics[scale=.45]{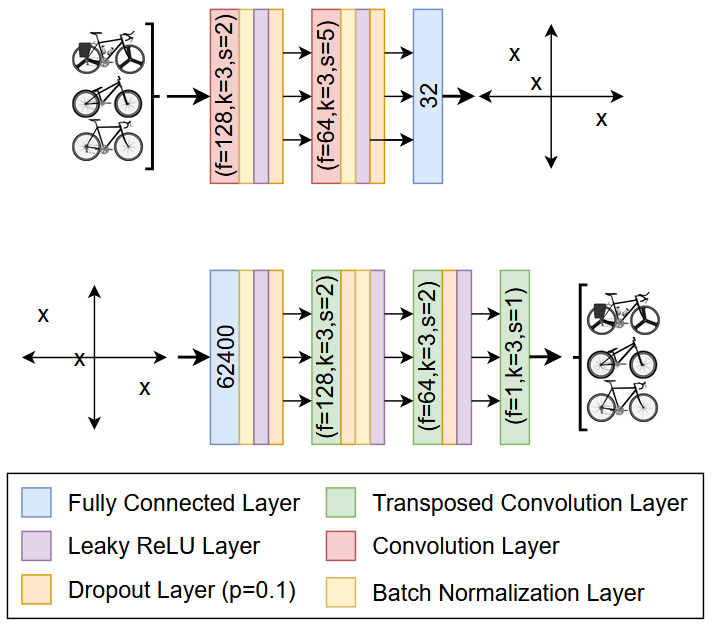}
    \caption{Model architecture of Variational Autoencoder for images.}
    \label{fig:VAE}
\end{figure}

\begin{figure}[h]
    \centering
    \includegraphics[scale=.37]{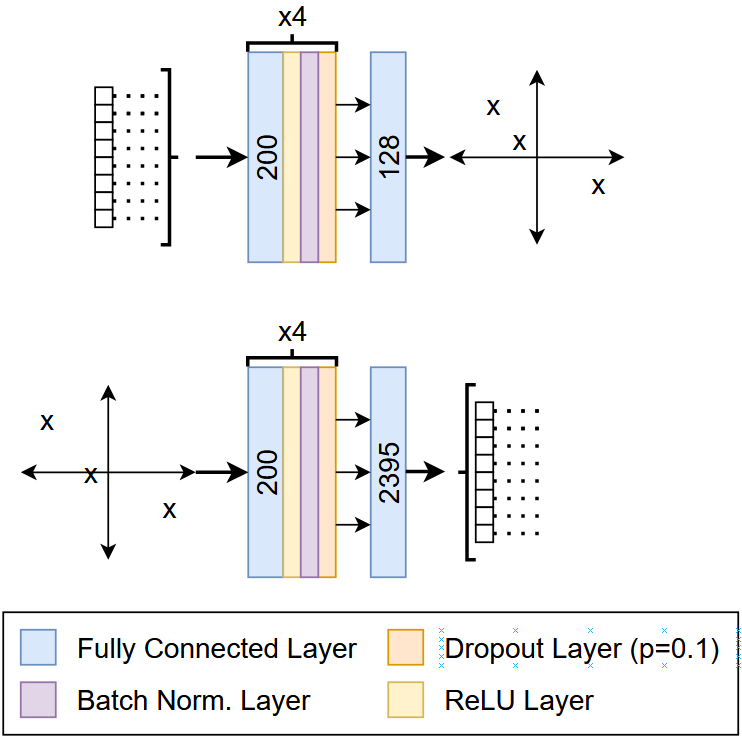}
    \caption{Model architecture of Variational Autoencoder for parametric data.}
    \label{fig:VAEd}
\end{figure}

\begin{figure}[h]
    \centering
    \includegraphics[scale=.45]{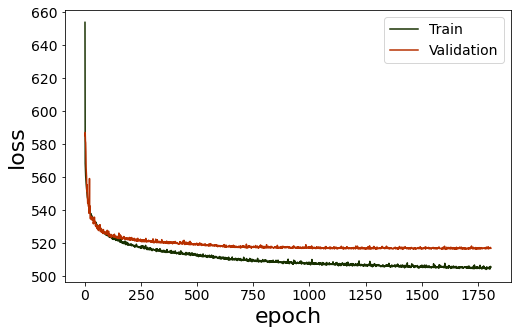}
    \caption{Training of Parametric VAE}
    \label{fig:datvaetrain}
\end{figure}

\begin{figure}[h]
    \centering
    \includegraphics[scale=.45]{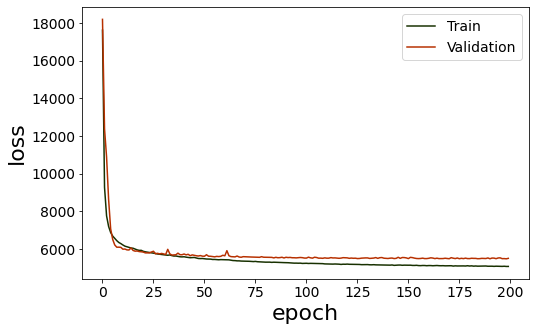}
    \caption{Training of Image VAE}
    \label{fig:imvaetrain}
\end{figure}

Figures~\ref{fig:VAE} and~\ref{fig:VAEd} show the architecture of the image VAE and parametric VAE respectively using the same labeling scheme as the classifiers in Appendix B1. The top portion of each diagram represents the encoder that maps a design to a latent space vector. The bottom represents the decoder that maps a latent vector to a design. Models were trained using the Adam optimizer~\cite{kingma2014adam} with a learning rate of $10^{-3}$ for the parametric VAE and $10^{-4}$ for the image VAE. Training is halted after 200 consecutive training epochs with no improvement for the parametric VAE and after 20 for the image VAE. Data dimensionality and train, validation, and test set splits are identical to the classifiers described in B2. The training was performed on a computer with Ryzen 9 5950x, 32 GB RAM, and Nvidia RTX3080 GPU. 
Training plots of the image and data VAEs are included in Figures~\ref{fig:datvaetrain} and \ref{fig:imvaetrain}. 

\section*{Appendix C2: Interpolation Setup}\label{appeninterp}
We define the interpolation and extrapolation vectors ($i_n$) and ($e_n$), respectively in terms of two original design vectors $m_1$ and $m_2$ as follows:
\begin{equation}
    i_k= m_1+k(m_2-m_1)*s_i,\,\,\, \forall\,\,\, k\in [1,n_i]
\end{equation}
\begin{equation}
    e_k= m_2+k(m_2-m_1)*s_e,\,\,\, \forall\,\,\, k\in [1,n_e]
\end{equation}
Here, $s_i$ and $s_e$ are the interpolation and extrapolation step sizes and $n_i$ and $n_e$ are the number of interpolation and extrapolation steps respectively. In this demonstration, we select $n_i=n_e=3$. In the case of interpolation, our step size is determined by the number of interpolation steps, while in extrapolation, we choose to select the step size such that our last extrapolation point is equally far from $m_2$ as the distance between $m_1$ and $m_2$:
\begin{equation}
    s_i=\frac{1}{n_i+1}
\end{equation}
\begin{equation}
    s_e=\frac{1}{n_e}
\end{equation}
Using this formulation, the results in Figure \ref{fig:InterExtra} are generated. 

\newpage

\section*{Appendix D1: Temporal Bias}\label{bias}
\begin{figure}[h]
    \centering
    \includegraphics[scale=.48]{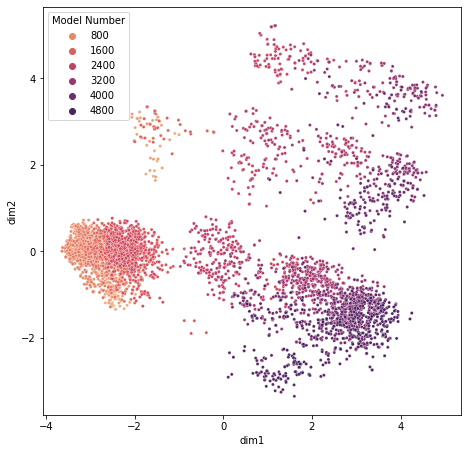}
    \caption{PCA embedding with designs labeled by model number illustrating the temporal bias in the dataset}
    \label{fig:bias}
\end{figure}

Figure \ref{fig:bias} contains a 2-Dimensional PCA embedding with models colored by model number. Clearly, there is a strong bias in the dataset based on model number, which is caused by the "status quo" of bike parameters evolving as new versions of the software were developed. This bias would not be apparent had we shuffled the data, and we intentionally left the data unshuffled in case the ordering of models proves valuable in future research. 
\end{document}